\documentclass[format=acmlarge, screen, review=false, timestamp=false, 
natbib=true, nonacm]{acmart}
\usepackage{amsmath}
\usepackage{longtable}
\usepackage{multirow}
\usepackage{subcaption}
\usepackage{ragged2e}
\usepackage{fontenc}
\usepackage{svg}
\usepackage[utf8]{inputenc}
\usepackage{booktabs}
\usepackage{url}
\usepackage{siunitx}
\usepackage[nocfg, nomentbl]{nomencl}

\setcitestyle{nosort}
\definecolor{t1_red}{RGB}{212,0,0}
\definecolor{t1_green}{RGB}{0,128,0}
\definecolor{t1_peach}{RGB}{255,204,170}
\definecolor{pastel_peach}{RGB}{255,230,213}
\definecolor{t1_light_green}{RGB}{170,222,135}
\definecolor{sky_blue}{RGB}{215,238,244}

\usepackage{tikz}
\usepackage{pgfplots}
\pgfplotsset{compat=1.18}
\usetikzlibrary{shapes.geometric, 
arrows.meta, positioning, calc}
\usepackage{svg}

\tikzstyle{startstop} = [rectangle, rounded corners, minimum width=2.5cm, minimum height=1cm,text centered, draw=black, fill=gray!20]
\tikzstyle{process} = [rectangle, minimum width=2.5cm, minimum height=1cm, text centered, draw=black, fill=white]
\tikzstyle{decision} = [diamond, aspect=2, minimum height=1.2cm, text centered, draw=black, fill=white]
\tikzstyle{datastore} = [cylinder, shape border rotate=90, minimum height=1.5cm, text centered, draw=black, aspect=0.25]
\tikzstyle{data} = [rectangle, minimum width=2.5cm, minimum height=1cm, text centered, draw=black, fill=white]
\tikzstyle{arrow} = [thick,->,>=stealth]
\tikzstyle{optional_arrow} = [thick,->,>=stealth, dashed]

\usepackage{dirtree}
\title[MECHBench]{MECHBench: A Set of Black-Box Optimization Benchmarks originated from Structural Mechanics
}
\author{Iván Olarte Rodríguez}
\orcid{0009-0005-0748-9069}
\affiliation{
\institution{Leiden University}
\department{LIACS}
\city{Leiden}
\country{The Netherlands}
}
\email{i.olarte.rodriguez@liacs.leidenuniv.nl}
\author{Maria Laura Santoni}
\orcid{0009-0007-7389-2555}
\affiliation{
\institution{Sorbonne Université, CNRS, LIP6}
\city{Paris}
\country{France}
}
\email{maria-laura.santoni@lip6.fr}
\author{Fabian Duddeck}
\orcid{0000-0001-8077-5014}
\affiliation{
\institution{Technical University of Munich}
\department{TUM School of Engineering \& Design}
\city{Munich}
\country{Germany}
}
\email{duddeck@tum.de}
\author{Carola Doerr}
\orcid{0000-0002-4981-3227}
\affiliation{
\institution{Sorbonne Université, CNRS, LIP6}
\city{Paris}
\country{France}
}
\email{carola.doerr@lip6.fr}
\author{Thomas Bäck}
\orcid{0000-0001-6768-1478}
\affiliation{
\institution{Leiden University}
\department{LIACS}
\city{Leiden}
\country{The Netherlands}
}
\email{t.h.w.baeck@liacs.leidenuniv.nl}
\author{Elena Raponi}
\orcid{0000-0001-6841-7409}
\affiliation{
\institution{Leiden University}
\department{LIACS}
\city{Leiden}
\country{The Netherlands}
}
\email{e.raponi@liacs.leidenuniv.nl}
\makenomenclature
\begin{document}

\begin{abstract}
Benchmarking is essential for developing and evaluating black-box optimization algorithms, providing a structured means to analyze their search behavior. Its effectiveness relies on carefully selected problem sets used for evaluation. To date, most established benchmark suites for black-box optimization consist of abstract or synthetic problems that only partially capture the complexities of real-world engineering applications, thereby severely limiting the insights that can be gained for application-oriented optimization scenarios and reducing their practical impact. To close this gap, we propose a new benchmarking suite that addresses it by presenting a curated set of optimization benchmarks rooted in structural mechanics. The current implemented benchmarks are derived from vehicle crashworthiness scenarios, which inherently require the use of gradient-free algorithms due to the non-smooth, highly non-linear nature of the underlying models. Within this paper, the reader will find descriptions of the physical context of each case, the corresponding optimization problem formulations, and clear guidelines on how to employ the suite.
\end{abstract}

\maketitle
\renewcommand{\shortauthors}{Olarte Rodriguez et al.}

\printnomenclature

\section{Introduction}
\label{sec:intro}
Optimization in structural mechanics plays a critical role in the design of lightweight, cost-efficient, and crash-resistant structures. In specific contexts such as vehicle crashworthiness, where occupant safety must be balanced with structural performance and weight constraints, optimization techniques are essential for developing effective energy-absorbing designs~\cite{dubois_vehicle_2004}.

Many applications in structural mechanics are modeled as non-linear dynamical systems due to phenomena such as fracture, damage, and material plasticity, which inherently exhibit non-linear behavior. This hinders the use of gradient-based optimizers since this implies that computation of sensitivities of the objectives with respect to each design variable is rather inefficient as small step sizes must be used \cite{fang_design_2017}. Furthermore, works like \cite{zabaras_continuum_2003} showed that in cases of highly nonlinear systems, the estimation of sensitivities via Finite Differences are uninformative due to noise.

Due to the limitations of gradient-based methods in the aforementioned applications, gradient-free optimization algorithms have gained significant attention. Notably, \citet{duddeck_multidisciplinary_2008} demonstrated that population-based gradient-free optimizers, when coupled with simulations of frontal crash scenarios using the Explicit Finite Element Method (EFEM), can achieve promising results. Nevertheless, the adoption of these methods remains a topic of ongoing debate.

On one hand, gradient-free algorithms typically require a large number of objective function evaluations. Despite advances in computational resources, conducting numerous high-fidelity crash simulations remains computationally expensive and time-consuming. Additionally, many sampled designs may be unfeasible or exhibit poor performance~\cite{fang_design_2017}.

On the other hand, standard gradient-free methods offer no theoretical guarantee of finding globally optimal designs, and their effectiveness depends on the problem at hand \cite{audet_derivative-free_2017,rios_derivative-free_2013}. This observation is consistent with the No Free Lunch (NFL) theorems for optimization, which state that no algorithm can universally outperform others across all problem classes \cite{wolpert_no_1997}. Consequently, the selection of an appropriate black-box optimization strategy becomes a critical challenge on top of a simulation or experimental pipeline.

The performance of black-box optimizers is typically assessed through experiments conducted on standardized benchmarking suites \cite{kudela_critical_2022}. Among the most widely used are COmparing COntinuous Optimizers (\texttt{COCO}) \cite{hansen_coco_2021} and \texttt{CEC} Benchmarking Suites \cite{CEC_Michalewicz2000}, which offer collections of synthetic test functions specifically designed for scalability. These benchmarks provide multiple instances by systematically shifting the location of the global optimum and modifying the optimization landscape using transformations such as rotations. A key feature of these benchmarks is their computational efficiency: objective evaluations are fast and deterministic, allowing extensive experimentation across multiple runs. This makes them a convenient and reproducible testbed for comparing black-box optimization algorithms.

However, an ongoing debate concerns whether strong performance on such synthetic benchmarks reliably translates to success on real-world expensive optimization problems. As highlighted by \citet{bliek_benchmarking_2023}, real-world problems often exhibit significantly more complex and irregular landscapes that are not well captured by simple mathematical functions, making the generalization of benchmark-based performance uncertain. In the domain of multi-objective optimization, a best-of-both-worlds approach has been pursued, where simplified and empirical models are used to design physical systems, as in the suites developed by \citet{jain_evolutionary_2014} and \citet{tanabe_easy_to_use_2020}. Nevertheless, this comes at the cost of reduced precision in evaluating the objectives. Moreover, simplified models are typically valid only within a limited range of the design variables, which also hinders global applicability.

\begin{sloppypar}
In turn, several benchmarking suites such as \texttt{EXPOBench} \cite{bliek_benchmarking_2023}, \texttt{SUMO} \cite{gorissen_surrogate_2010}, \texttt{DAFoam}~\cite{he_aerodynamic_2018,he_dafoam_2020}, and the \texttt{CFD} \texttt{Optimization} \texttt{Suite} \cite{daniels_suite_2018} feature expensive, real-world optimization problems involving simulation tools like \texttt{OpenFOAM}. However, these frameworks do not include problems derived from structural mechanics.
\end{sloppypar}

In contrast, the \texttt{MOPTA08} benchmark~\cite{jones2008large}, which is directly inspired by vehicle crashworthiness optimization, is based on a surrogate model trained on a large set of high-fidelity finite element simulations. While this approach eliminates the cost of online simulations, it suffers from limited scalability, as its formulation is fixed to a 124-dimensional design space. A similar limitation exists in the \texttt{Mazda} \texttt{CdMOBP} benchmark~\cite{kohira_proposal_2018}; however, in this case, all design variables are discrete. Furthermore, an issue with these aforementioned benchmarks revolves around the complexity of setting up the cases and the simulation environments. In our case, we drew inspiration from the \texttt{IOHExperimenter} \cite{IOHexperimenter} interface, ensuring that our problems can be called in a straightforward and standardized way. This design choice not only reduces setup complexity but also makes it easy to embed the benchmark problems into other codebases by following the structure from \citet{doerr_iohprofiler_2018} for function calling.

To address the gap between structural designers seeking effective optimization tools and researchers developing black-box optimization algorithms in need of realistic test environments, we introduce \texttt{MECHBench}. The benchmark focuses on three representative problems designed to enable consistent and reproducible evaluation of algorithms in terms of objective quality, constraint satisfaction, convergence behavior, and computational efficiency. Each of those can generate a broad family of related optimization tasks. Every simulation produces multiple physical outputs that can be flexibly combined to define diverse objectives and constraints.

In summary, our \textbf{main contributions} are:

\begin{itemize}
    \item We introduce a suite of optimization problems motivated by structural mechanics, designed to capture key challenges such as high dimensionality, nonlinearity, and symmetries.
    \item To support rigorous and comparable evaluation, we establish standardized baselines and performance metrics
    \item All code and datasets are released openly to foster transparency and reproducibility within the optimization, machine learning and the engineering design communities. The benchmarks are publicly available at our GitHub Repository (\url{https://github.com/BayesOptApp/MECHBench}). 
\end{itemize}

\section{Implementation}
To enable consistent evaluation of optimization algorithms in structural design, we introduce a set of mechanical benchmarks based on simplified crash simulations. Each benchmark, represented by a problem instance within a standardized pipeline, is designed to decouple the simulation process from the optimization logic. This modular structure allows integration with a wide range of optimizers, regardless of the programming interface, including non-Python environments. This approach contrasts with earlier works by \citet{hunkeler_shape_2013} and \citet{Volz2007}, in which the optimization loop was tightly integrated into a commercial software, thereby limiting flexibility and external interfacing.

\subsection{Benchmark Workflow}
The benchmarking process is illustrated in Figure \ref{fig:proposed_optim_loop} and consists of the following key stages:

\begin{itemize}
    \item \textsc{Design Variable Update}:
The optimization algorithm proposes a new configuration by modifying a set of $d$ design variables (DVs). All the cases analyzed in these benchmarks correspond to shape optimization; thus, the design variables only affect the geometry of the underlying structure.
\item \textsc{Model Assembly}: The updated input data is used to assemble a new finite element (FE) model. This includes integrating the DV changes into the model structure and preparing it for simulation. 
\item \textsc{Simulation Execution}: The assembled FE model is simulated using OpenRadioss \cite{openradioss2025}, which is an open-source simulation solver based on EFEM formulation. This step generates detailed time-history and summary results characterizing the crash response. 
\item \textsc{Post-Processing and Objective Evaluation}: Simulation output files, particularly the time histories in \texttt{*.csv} format, are automatically parsed to extract relevant variables to compute the objectives. Furthermore, for visual inspection and further post-processing, the user can set an option to generate \texttt{*.vtk} files for different simulation times.
\end{itemize}
\begin{figure}[htbp]
    \centering
    \resizebox{\textwidth}{!}{
\begin{tikzpicture}[node distance=1.25cm and 1.25cm]


\node (optimizer) [datastore, 
					minimum height=2cm, 
					fill=gray!30,
					 align=center] {Optimization Algorithm\\(Solver)};

\node (decision) [decision, above=of optimizer] {Termination Satisfied?};


\node (end) [startstop, left=2.5cm of decision] {End};
\node (modDV) [startstop, right=2.5cm of decision] {Modified DV};

\node (probinstance) [datastore, minimum height=1.9cm, fill=red!20, right=of modDV] {Problem};

\node (femodel) [startstop, fill=red!30, below=of probinstance] {FE model};
\node (assembly) [startstop, fill=blue!20, below=of femodel] {Model Assembly};

\node (solver) [datastore, 
				minimum height=2.2cm, 
				fill=green!20, 
				below=of assembly,
				align=center] {EFEM Simulation\\ via OpenRadioss};
\node (results) [startstop, fill=green!30, left=of solver] {Results};
\node (th) [data, above=of results] {Time History};
\node (anim) [data, left=of th] {Animation Files (\texttt{*.vtk})};
\node (objectives) [data, left=of femodel, above=of th, right=of optimizer] {Objective values};

\draw [arrow] (optimizer) -- (decision);
\draw [arrow] (decision) -- node[above]{Yes} (end);
\draw [arrow] (decision) -- node[above]{No} (modDV);
\draw [arrow] (modDV) -- (probinstance);
\draw [arrow] (probinstance) -- (femodel);
\draw [arrow] (femodel) -- (assembly);
\draw [arrow] (assembly) -- (solver);
\draw [arrow] (solver) -- (results);
\draw [arrow] (results) -- (th);
\draw [arrow] (objectives) -- (optimizer);
\draw [arrow] (th) -- (objectives);
\draw[optional_arrow] (results) -- (anim);

\end{tikzpicture}
}
    \caption[Proposed optimization loop.]{Proposed optimization loop. To initiate the loop, the optimization algorithm, objective values, and problem number must be defined \textit{a priori}. The blocks connected with solid arrows represent the necessary steps of the loop, whereas the dashed arrows indicate optional steps or processes that do not interrupt the loop. Adapted from \cite{hunkeler_shape_2013, Volz2007}.}
    \label{fig:proposed_optim_loop}
\end{figure}
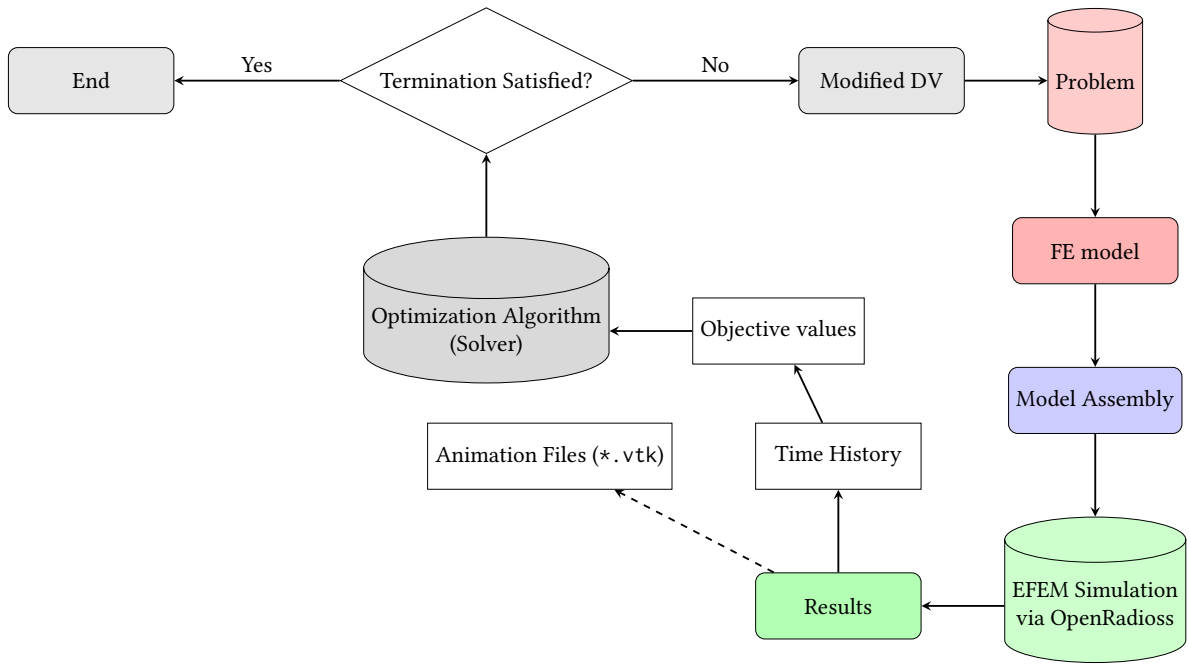
\begin{figure}[htbp!]
	\begin{footnotesize}

    \begin{minipage}{0.4\linewidth} 
	\dirtree{%
	  .1 Project Root.
	  .2 tests.py.
	  .2 main.py.
	  .2 results.
	  .2 src.
	  .3 sob.
	  .4 \_\_init\_\_.py.
	  .4 problems.py.
	  .4 fem.py.
	  .4 mesh.py.
	  .4 solver.py.
	  .4 utils.
	  .5 run\_openradioss.py.
	  .5 solver\_setup.py.
	  .4 lib.
	  .5 gmsh\_base\_meshes.py.
	  .5 starbox\_gmsh.py.
	  .5 crashtube\_gmsh.py.
	  .5 three\_point\_bending\_gmsh.py.
	  .5 ThreePointBending\_0001.rad.
	}
	\end{minipage}
\end{footnotesize}

	\Description{The project is organized as follows. At the top level, the Project Root contains
the main entry point (main.py), a test script (tests.py), and a results directory
which shows sample files generated after a generated simulation. The main source code is located under src/, where the sob/
subpackage holds the core functionality. This includes modules for problem
definitions (problems.py), finite element routines (fem.py), mesh handling (mesh.py),
solvers (solver.py), and post-processing tools (post_processing.py).

Within the second level or namely sob/, there are two utility subdirectories. The utils/ folder provides
helper scripts such as run_openradioss.py for starting the subprocesses  and solver_setup.py
for preparing solver configurations. Finally, the third level or the lib/ folder contains geometry and mesh
generation scripts, including gmsh_base_meshes.py as a foundation, and specialized
generators like starbox_gmsh.py, crashtube_gmsh.py, and three_point_bending_gmsh.py.
}
    \caption[\texttt{MECHBench} directory structure.]{\texttt{MECHBench} directory structure. Currently, the project contains 3 levels; Level 1 is where the \texttt{main.py} should go. On the same level, the repository has the folder \texttt{results} which contain sample simulation-generated files. The main source code is located under \texttt{src/}, where the \texttt{sob/subpackage} holds the core functionality. This includes modules for problem definitions (\texttt{problems.py}), finite element routines (\texttt{fem.py}), mesh handling (\texttt{mesh.py}), and solvers (\texttt{solver.py}). Within the second level or \texttt{sob/}, there are two utility subdirectories. The \texttt{utils/} folder provides helper scripts such as \texttt{run\_openradioss.py} for starting the subprocesses and \texttt{solver\_setup.py} for preparing solver configurations. Finally, the third level or \texttt{lib/} folder contains geometry and mesh generation scripts, including \texttt{gmsh\_base\_meshes.py} as a foundation, and specialized generators like \texttt{starbox\_gmsh.py}, \texttt{crashtube\_gmsh.py}, and \texttt{three\_point\_bending\_gmsh.py.}  }
    \label{fig:dir_structure}
\end{figure}
\subsection{System Design Principles}
The benchmark system is designed with the following principles:
\begin{itemize}
    \item \textsc{Modularity}: Each stage of the workflow is modularized to allow for easy substitution or extension (e.g., changing the FE solver or the objective functions).
    \item \textsc{Separation of concerns}: The optimization algorithm and the crash simulation pipeline operate independently and communicate through a clear interface. This facilitates the integration of new optimizers without modifying the core benchmark logic.
    \item \textsc{Open-Source}: All steps in the pipeline, except for the optimizer, rely on open-source packages (such as \texttt{NumPy}~\cite{harris2020array} and \texttt{Gmsh}~\cite{geuzaine_gmsh_2009}) and produce outputs in open file formats, eliminating the need for proprietary software.
    \item \textsc{Execution Scalability}: The framework supports execution on local machines or high-performance computing (HPC) clusters.
\end{itemize}
\subsection{Usage}
Although Section \ref{sec:benchmarks} presents the physical bounds of the design variables, these bounds are not directly imposed in the optimizer. Instead, each design variable is mapped to the normalized domain $[-5,5]^d$, consistent with the \texttt{COCO} benchmarks.

The interface is mainly built in \texttt{Python} and it works in a sequential fashion, as each problem object processes design variables vectors one at a time. This is because of the arrangement of files that OpenRadioss generates such that these shall be stored in a distinctive directory.
Detailed instructions are available in our GitHub repository. For completeness, the usage and setup are outlined below in three steps:

\begin{enumerate}
    \item \textsc{Clone the repository}: The structure of the repository must follow the one shown in Figure \ref{fig:dir_structure}.
   \item \textsc{Download or build the OpenRadioss binaries}: Precompiled OpenRadioss binaries are available for both Windows and Linux systems from the official website. Alternatively, you can build the binaries from the source repository, though the resulting executables are functionally equivalent to the prebuilt versions. The repository is actively maintained by Altair, the company behind OpenRadioss, along with a dedicated community of contributors.
   \item \textsc{Set the case}: The problem object from Figure \ref{fig:proposed_optim_loop} needs four ingredients, namely: i) The path to the OpenRadioss binaries, ii) The problem number (identified from 1 to 3), iii) the problem dimension to evaluate, and iv) the list of objectives to extract. After this setting, the user might start testing different design variable vectors.
\end{enumerate}
To maintain modularity in the benchmarks, users can either call problem-specific objective functions (based on those presented in Section \ref{sec:benchmarks}) or define tailor-made problems by extracting energy-based crashworthiness metrics, such as those proposed by \citet{fang_design_2017}.
\section{Benchmark Problems}
\label{sec:benchmarks}
The numerical benchmark problems introduced in this section are constructed to rigorously evaluate optimization algorithms in the context of nonlinear crashworthiness analysis. Each benchmark comprises a structural component subjected to dynamic impact loading, where the design task involves adjusting geometric parameters such as thickness distributions, trigger configurations or rib dimensions, with the aim of enhancing energy absorption capacity, reducing peak force levels, or minimizing structural mass while respecting prescribed deformation limits.

To provide practitioners with practical guidance regarding computational requirements, we evaluated the wall-clock time associated with a single OpenRadioss simulation for each benchmark problem under varying levels of parallelization. The runtime of an individual simulation is not influenced by the dimensionality of the optimization problem, as the finite element model and mesh are fixed once defined. However, it does depend on the specific design variable configuration, since the elemental time step enforced by OpenRadioss is sensitive to geometric features and local stiffness variations dictated by the chosen parameters.

Table~\ref{tab:runtime} summarizes the average runtime per simulation, together with the corresponding number of CPU cores utilized (both before and after the OpenRadioss call), for all three benchmark cases. All computations were carried out on the Academic Leiden Interdisciplinary Cluster Environment (ALICE) at Leiden University, exclusively employing Intel Xeon 6126 processors operating at 2.60 \unit{GHz}. Each compute node is equipped with 24 cores and 384 \unit{GB} of RAM.
\begin{table}[htbp!]
\centering
\caption{Average wall-clock time (in seconds) per simulation, across 45 evaluations, for each of the three problems.}
\label{tab:runtime}
\resizebox{\textwidth}{!}{%
\begin{tabular}{cl|cccc|}
\cline{3-6}
\multicolumn{1}{l}{} &  & \multicolumn{4}{c|}{\textbf{Number of Cores}}                                                                    \\ \cline{3-6} 
\multicolumn{1}{l}{} &  & \multicolumn{1}{c|}{\textit{1}} & \multicolumn{1}{c|}{\textit{2}} & \multicolumn{1}{c|}{\textit{4}} & \textit{8} \\ \hline
\multicolumn{1}{|c|}{\multirow{3}{*}{\textbf{Problem}}} &
  \textit{Star Shaped Crash-Box} &
  \multicolumn{1}{c|}{$842.5 \pm 19.1$} &
  \multicolumn{1}{c|}{$757.9 \pm 16.8$} &
  \multicolumn{1}{c|}{$481.3 \pm 12.4$} &
  $328.5 \pm 8.1$ \\ \cline{2-6} 
\multicolumn{1}{|c|}{} &
  \textit{Three Point Bending of Layered Beam} &
  \multicolumn{1}{c|}{$234.3 \pm 1.6$} &
  \multicolumn{1}{c|}{$203.6 \pm 1.3$} &
  \multicolumn{1}{c|}{$117.4 \pm 1.0$} &
  $71.7 \pm 0.5$ \\ \cline{2-6} 
\multicolumn{1}{|c|}{} &
  \textit{Long Crash Tube} &
  \multicolumn{1}{c|}{$1918.5 \pm 16.0$} &
  \multicolumn{1}{c|}{$1532.4 \pm 12.6$} &
  \multicolumn{1}{c|}{$858.6 \pm 9.6$} &
  $489.2 \pm 5.3$ \\ \hline
\end{tabular}%
}
\end{table}
\subsection{Star Shaped Crash-Box}
\subsubsection{Problem Description}
\begin{figure}[htbp]
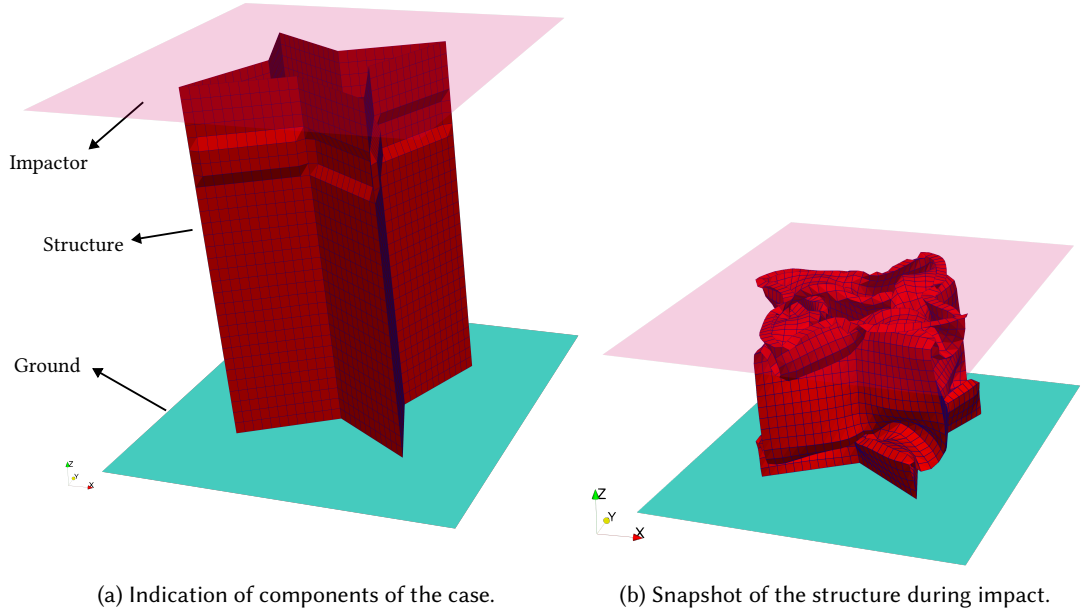

    \centering
    \begin{subfigure}[b]{0.48\linewidth}
    	\centering
	    \begin{footnotesize}
	    	\includesvg[width=\textwidth]{SVGFigures/starbox_diagram_with_arrows.svg}
	    \end{footnotesize}
	    \caption{Indication of components of the case.}
	    \label{fig:starbox_representation_1}
    \end{subfigure}
    \begin{subfigure}[b]{0.41\linewidth}
	    \centering
	    \includesvg[width=\textwidth]{SVGFigures/starbox_diagram_crashed.svg}
	    \caption{Snapshot of the structure during impact.}
	    \label{fig:starbox_representation_2}
    \end{subfigure}
    \caption{Star-shaped crash-box case. The structure is subdivided into smaller connected parts, forming a finite element mesh shown by the \textcolor{blue}{blue} lines. In the \nomenclature{$z-$}{$z$ position in $\mathbb{R}^{3}$}{\unit{mm}}{}$z$-direction, the mesh is arranged in 30 rows of elements, wherein each row is assigned the same thickness value $t_h$.}
    \label{fig:starbox_representation}
\end{figure}

The objective in this case is to optimize both the thickness profile and the transverse profile of the crash box shown in Figure \ref{fig:starbox_representation}. In this configuration, the structure is clamped to the ground at its base, preventing any rotation or displacement, while an impactor moves toward the crash box from above. The goal is to maximize the specific energy absorption (SEA), subject to an intrusion constraint of 60 mm, corresponding to half the extrusion length of the crash box. This formulation is inspired by the work of \citet{hunkeler_shape_2013}, with the key difference that the constraint here is defined in terms of maximum intrusion rather than peak crash force.

Put simply, the design must absorb as much crash energy as possible while preventing the structure from being compressed by more than 60 \unit{mm}. The goal is to ensure efficient energy dissipation during impact while maintaining a deformation limit that preserves the required structural clearance.

\subsubsection{Variable Definition}
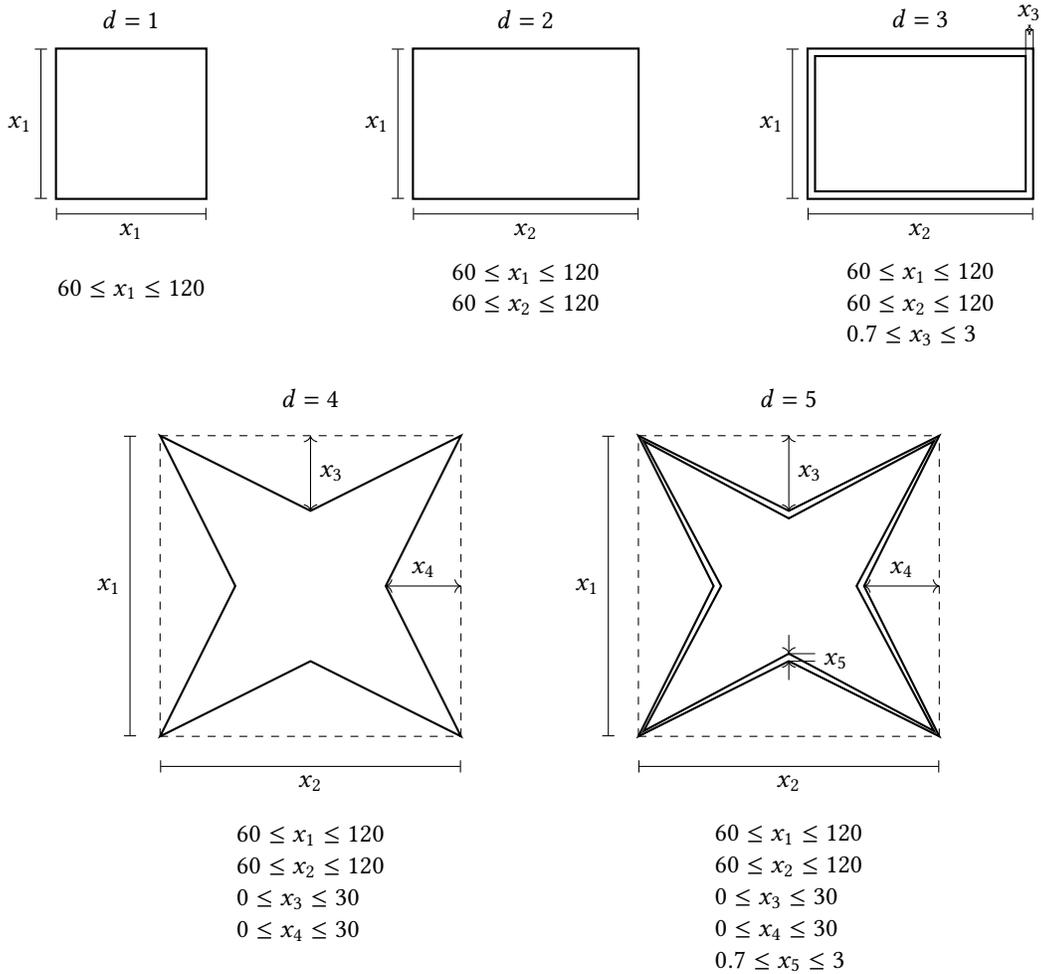
\begin{figure}[htbp]
    \centering
    \begin{tikzpicture}[scale=1.0]

\begin{scope}
	\node at (0,1.4) {$d=1$};
	\coordinate (A1) at (1,1);
	\coordinate (A2) at (-1,1);
	\coordinate (A3) at (-1,-1);
	\coordinate (A4) at (1,-1);
	\draw[thick] (A1)--(A2)--(A3)--(A4)--cycle;
	
	\draw[|-|] (-1.2,-1) -- (-1.2,1) node[midway,left] {$x_{1}$};
	\draw[|-|] (-1.,-1.2) -- (1.,-1.2) node[midway,below] {$x_{1}$};
	\node[align=left] at (0,-2.2) {$60 \leq x_1 \leq 120$};
\end{scope}

\begin{scope}[xshift=0.33\linewidth]
	\node at (0,1.4) {$d=2$};
	\coordinate (A1) at (1.5,1);
	\coordinate (A2) at (-1.5,1);
	\coordinate (A3) at (-1.5,-1);
	\coordinate (A4) at (1.5,-1);
	\draw[thick] (A1)--(A2)--(A3)--(A4)--cycle;
	
	\draw[|-|] (-1.7,-1) -- (-1.7,1) node[midway,left] {$x_{1}$};
	\draw[|-|] (-1.5,-1.2) -- (1.5,-1.2) node[midway,below] {$x_{2}$};
	\node[align=left] at (0,-2.2) {$60 \leq x_1 \leq 120$ \\ $60 \leq x_2 \leq 120 $};
\end{scope}

\begin{scope}[xshift=0.66\linewidth]
	\node at (0,1.4) {$d=3$};
	\coordinate (A1) at (1.5,1);
	\coordinate (A2) at (-1.5,1);
	\coordinate (A3) at (-1.5,-1);
	\coordinate (A4) at (1.5,-1);
	\draw[thick] (A1)--(A2)--(A3)--(A4)--cycle;
	
	\coordinate (A5) at (1.4,0.9);
	\coordinate (A6) at (-1.4,0.9);
	\coordinate (A7) at (-1.4,-0.9);
	\coordinate (A8) at (1.4,-0.9);
	\draw[thick] (A5)--(A6)--(A7)--(A8)--cycle;
	
	\draw[|-|] (-1.7,-1) -- (-1.7,1) node[midway,left] {$x_{1}$};
	\draw[|-|] (-1.5,-1.2) -- (1.5,-1.2) node[midway,below] {$x_{2}$};
	
	\draw[very thin] (1.4,0.9) -- (1.4,1.25);
	\draw[very thin] (1.5,0.9) -- (1.5,1.25);
	\draw[<->, ultra thin] (1.4,1.25) -- (1.5,1.25) node[midway,above] {$x_{3}$};
	\node[align=left] at (0,-2.4) {$60 \leq x_1 \leq 120$ \\ $60 \leq x_2 \leq 120 $ \\ $ 0.7 \leq x_3 \leq 3 $};
\end{scope}

\begin{scope}[xshift=0.15\linewidth, yshift=-175pt]
	\coordinate (A1) at (2,2);
	\coordinate (A2) at (0,1);
	\coordinate (A3) at (-2,2);
	\coordinate (A4) at (-1,0);
	\coordinate (A5) at (-2,-2);
	\coordinate (A6) at (0,-1);
	\coordinate (A7) at (2,-2);
	\coordinate (A8) at (1,0);

\draw[thick] (A1)--(A2)--(A3)--(A4)--(A5)--(A6)--(A7)--(A8)--cycle;

\draw[dashed] (-2,2) -- (2,2);
\draw[dashed] (-2,-2) -- (2,-2);
\draw[dashed] (-2,-2) -- (-2,2);
\draw[dashed] (2,-2) -- (2,2);

\draw[|-|] (-2.4,-2) -- (-2.4,2) node[midway,left] {$x_1$};

\draw[|-|] (-2,-2.4) -- (2,-2.4) node[midway,below] {$x_2$};

\draw[<->] (0,2) -- (0,1) node[midway,right] {$x_3$};
\draw[<->] (1,0) -- (2,0) node[midway,above] {$x_4$};

\node at (0,2.5) {$d=4$};

\node[align=left] at (0,-3.95) {$60 \leq x_1 \leq 120$ \\ $60 \leq x_2 \leq 120 $ \\ $ 0 \leq x_3 \leq 30 $ \\ $ 0 \leq x_4 \leq 30 $};

\end{scope}

\begin{scope}[xshift=0.55\linewidth,yshift=-175pt]
	\coordinate (A1) at (2.0000,2.0000);
	\coordinate (A2) at (0.0000,1.0000);
	\coordinate (A3) at (-2.0000,2.0000);
	\coordinate (A4) at (-1.0000,0.0000);
	\coordinate (A5) at (-2.0000,-2.0000);
	\coordinate (A6) at (0.0000,-1.0000);
	\coordinate (A7) at (2.0000,-2.0000);
	\coordinate (A8) at (1.0000,0.0000);
	
	\coordinate (A9) at (1.9293,1.9293);
	\coordinate (A10) at (0.0000,0.9000);
	\coordinate (A11) at (-1.9293,1.9293);
	\coordinate (A12) at (-0.9000,0.0000);
	\coordinate (A13) at (-1.9293,-1.9293);
	\coordinate (A14) at (0.0000,-0.9000);
	\coordinate (A15) at (1.9293,-1.9293);
	\coordinate (A16) at (0.9000,0.0000);

\draw[thick] (A1)--(A2)--(A3)--(A4)--(A5)--(A6)--(A7)--(A8)--cycle;
\draw[thick] (A9)--(A10)--(A11)--(A12)--(A13)--(A14)--(A15)--(A16)--cycle;

\draw[dashed] (-2,2) -- (2,2);
\draw[dashed] (-2,-2) -- (2,-2);
\draw[dashed] (-2,-2) -- (-2,2);
\draw[dashed] (2,-2) -- (2,2);

\draw[|-|] (-2.4,-2) -- (-2.4,2) node[midway,left] {$x_1$};

\draw[|-|] (-2,-2.4) -- (2,-2.4) node[midway,below] {$x_2$};

\draw[<->] (0,2) -- (0,1) node[midway,right] {$x_3$};
\draw[<->] (1,0) -- (2,0) node[midway,above] {$x_4$};

\draw[->] (0,-0.65) -- (0,-0.9) ;
\draw[<-] (0,-1) -- (0,-1.25) ;

\draw[-, very thin] (0,-0.9) -- (0.35,-0.9) ;
\draw[-, very thin] (0,-1.0) -- (0.35,-1.0)  node[right] {$x_5$} ;

\node at (0,2.5) {$d=5$};

\node[align=left] at (0,-4.15) {$60 \leq x_1 \leq 120$ \\ $60 \leq x_2 \leq 120 $ \\ $ 0 \leq x_3 \leq 30 $ \\ $ 0 \leq x_4 \leq 30 $ \\ $0.7 \leq x_5 \leq 3$};

\end{scope}

\end{tikzpicture}
    \caption{Optimization problem parameterization for $d\leq5$.  The ranges of each variable are expressed in mm. The drawings are not in scale. For the cases where thickness is not controlled such as the cases where $d=\{1,2,4\}$, the wall thickness $t_h$ is set constant to 2.1 \unit{mm}.}
    \label{fig:Hunkeler_crash_box_base_case}
\end{figure}

\begin{figure}[htbp]
    \centering
	
	 \begin{tikzpicture}
\tikzset{
    every pin/.style={fill=yellow!50!white,rectangle,rounded corners=3pt,font=\tiny},
    small dot/.style={fill=black,circle,scale=0.3},
}
  \begin{axis}[
  	clip=false,
    axis x line=bottom,
    axis y line=left,
    grid=major,
    xlabel = $t_h$,
    ylabel = $z$,
    xmin = 0.7, xmax = 3.1,
    ymin = 0, ymax = 130,
    xtick={0.7,1.85,3.0},
    ytick={0,20,40,60,80,100,120},
    xminorgrids=true,
    width=0.4\linewidth,
    height=0.55\linewidth,
    minor x tick num=1,
    every axis y label/.style={
            at={(ticklabel cs:0.5)},anchor=center,
        },
    every axis x label/.style={
            at={(ticklabel cs:0.5)},anchor=north,
        },
     axis y line shift=60pt,
     axis x line shift=5pt,
    nodes near coords,
    every node near coord/.append style={anchor=west},
    point meta=explicit symbolic
  ]

    \addplot[
      color=blue,
      thick,
      mark=*,
    ]
    table[meta=label] {
      x    y    label
      1.6  0    $x_{5}$
      2.7 60    $x_{6}$
      0.95 120    $x_{7}$
    };
     \draw [black, thin, |-|] ([xshift=30pt] yticklabel* cs:0)  -- ([xshift=30pt] yticklabel* cs:0.461538) node[midway, anchor=west] {$60$};
     \draw [black,thin,-|]      ([xshift=30pt] yticklabel* cs:0.46153846153846153846153846153846)     -- ([xshift=30pt] yticklabel* cs:0.92307692307692307692307692307692) node[midway, anchor=west] {$60$};
     \draw [red, dashed, thick] (yticklabel* cs:0.43) node[left, anchor=east] {$\bar{z}$} -- (axis description cs:0.80,0.43);
     \draw [red, dashed, thick, ->] (axis description cs:0.80,0.43)   -- (xticklabel* cs:0.79) node[below, anchor=north] {$\bar{t_{h}}$};

  \end{axis}
  

%

\end{tikzpicture}
    \caption[Thickness distribution example for $d=7$.]{Thickness distribution example for $d=7$, wherein 3 control points are set at uniformly distributed heights between $0$ and $120$ \unit{mm} and each point has an assigned $t_h(z)$. The wall thickness $\bar{t_h}$  at any intermediate point $\bar{z}$ is computed via linear interpolation of assigned $t_{h}$ of the neighboring control points. This procedure is used to compute the wall thickness $t_h$ of a row of finite elements by getting the interpolated $\bar{t_h}$ at the element barycenter.}
    \label{fig:starbox_distributions}
\end{figure}
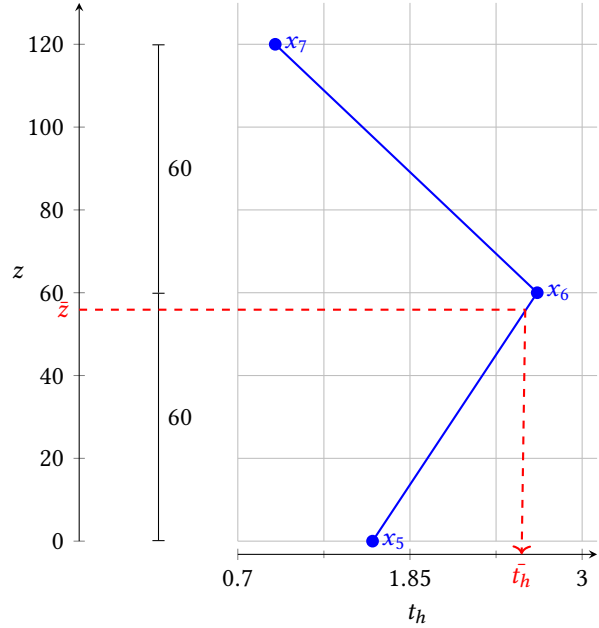

For a lower-dimensional setting, we directly retrieve the cases shown in \citet{hunkeler_shape_2013}, for \nomenclature{$d$}{Problem Dimension}{}{} $d \in \left\{2, 3, 4 ,5 \right\}$ thus we added a $d=1$ case by just defining a square. These settings are represented in Figure \ref{fig:Hunkeler_crash_box_base_case}.
For higher-dimensional settings ($6 \leq d \leq 34$), the definitions of $(x_1, x_2, x_3, x_4)$ remain identical to those in the four-dimensional case shown in Figure \ref{fig:Hunkeler_crash_box_base_case}. For indices $i$ ranging from 6 to 34, the variables define the control points of a piecewise-linear wall thickness profile of the crash box. An illustrative example for $d = 7$, corresponding to three control points, is provided in Figure \ref{fig:starbox_distributions}. For $d < 34$, the thickness of each of the 30 rows of elements is obtained by interpolating the distributions at the $z$-coordinate of the element barycenters. At $d = 34$, each row of the finite element mesh is assigned an independent thickness value, without interpolation. Consistent with the bounds in Figure \ref{fig:Hunkeler_crash_box_base_case}, all control points satisfy $0.7 \ \unit{mm} \leq x_i \leq 3 \ \unit{mm}$ for every index $i$ between 6 and 34.

\subsubsection{Standard Mathematical Formulation}

\begin{equation}
\begin{aligned}
\max_{\mathbf{x}} \;\; & \text{SEA}(\mathbf{x}) = & \frac{E_{\text{abs}}(\mathbf{x})}{m_{s}(\mathbf{x})} \\
\text{s.t.} \;\; & \delta(\mathbf{x})  \leq & 60 \ \unit{mm}. 
\end{aligned}
\label{eq:standard_problem_star_box}
\end{equation}%
\nomenclature{$\text{SEA}$}{Specific Energy Absorbed\nomrefeq}{J/kg}{}
\nomenclature{$E_{\text{abs}}$}{Absorbed Energy by structure\nomrefeq}{J}{}
\nomenclature{$m_s$}{Mass of the structure\nomrefeq}{kg}{}
\nomenclature{$\delta$}{Intrusion\nomrefeq}{mm}{}
\nomenclature{$\mathbf{x}$}{Input design/parameter vector}{}{}
With this objective, the structure is modified to maximize the specific energy absorption (SEA), defined as the ratio of the absorbed impact energy ($E_\text{abs}$), which is dissipated primarily through plastic deformation, to the total mass of the structure ($m_{s}$). This metric promotes designs that achieve high energy absorption while maintaining low structural weight.

\subsubsection{Reformulation}
Given that the problem is originally constrained, we formulated a new objective function which can be used with unconstrained optimizers. Contrary to \citet{raponi_global_2025}, who constructed a penalized objective function, we define a piecewise function instead as:
\begin{equation}
\min_{\mathbf{x}} \, \text{Penalized SEA}(\mathbf{x}) =
\begin{cases}
-\text{SEA}(\mathbf{x}), & \text{if } \delta(\mathbf{x}) \leq 60 \ \unit{mm}, \\
100 \, (\delta(\mathbf{x}) - 60), & \text{if } \delta(\mathbf{x}) > 60 \ \unit{mm}.
\end{cases}
\label{eq:reformulated_problem_star_box}
\end{equation}

With this modification, we specifically aim to avoid the need for assigning a de facto penalty weight through multiple preliminary runs. In the aforementioned case, such weight may still be insufficient to ensure that the maximum objective value within the feasible region exceeds that of the infeasible region, particularly when the latter corresponds to the lower bounds of wall thickness.

\subsection{Three Point Bending of Layered Beam}
\label{subsec:three_point_bending}
\subsubsection{Problem Description}
\begin{figure}[htbp]
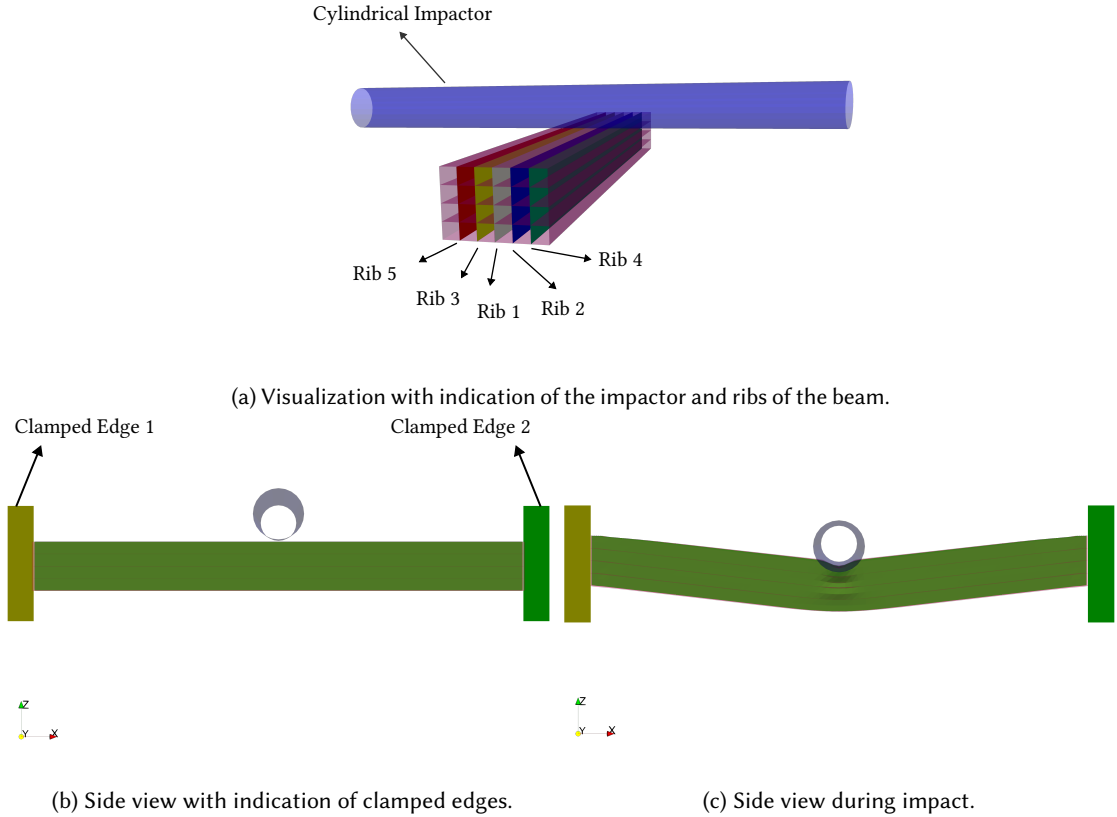

    \centering
    \begin{subfigure}[b]{0.55\textwidth}
	    \centering
	    \begin{footnotesize}
	    	\includesvg[width=\linewidth]{SVGFigures/three_point_bending_diagram_with_arrows.svg}
	    \end{footnotesize}
	    \caption{Visualization with indication of the impactor and ribs of the beam.}
	    \label{fig:three_point_bending_representation_1}
    \end{subfigure}
    \vfill
    \begin{subfigure}[b]{0.46\textwidth}
	    \begin{footnotesize}
	    	\includesvg[width=\linewidth]{SVGFigures/three_point_bending_diagram_with_arrows_2.svg}
	    \end{footnotesize}
	    \caption{Side view with indication of clamped edges.}
	    \label{fig:three_point_bending_representation_2}
    \end{subfigure}
    \begin{subfigure}[b]{0.46\textwidth}
	    \begin{footnotesize}
	    	\includesvg[width=\linewidth]{SVGFigures/three_point_bending_diagram_without_arrows.svg}
	    \end{footnotesize}
	    \caption{Side view during impact.}
	    \label{fig:three_point_bending_representation_3}
    \end{subfigure}
\caption{Representation of the Three Point Bending of Layered Beam case.}
\label{fig:three_point_bending_representation_overall}
\end{figure}

In this case, the beam consists of five ribs, as illustrated in Figure \ref{fig:three_point_bending_representation_1}. The objective is to reduce the overall weight of the structure by adjusting the thickness of each rib. During testing, the beam is fixed firmly at both ends (clamped supports), as shown in Figure \ref{fig:three_point_bending_representation_2}, so that it cannot move or rotate at the boundaries. This setup allows the ribs to deform under load in a controlled way, while the optimization seeks the lightest design that can still withstand the applied forces.  

A helpful analogy is to think of the ribs like the wooden slats of a bookshelf: making them thinner reduces the weight of the shelf, but if they are too thin, the shelf will bend or even break under load. The goal is therefore to find the right balance between lightness and strength.

We drew inspiration from the case presented by \citet{kaps_hierarchical_2022}, adopting the same mesh sizing and material properties described therein. Notably, \citet{kaps_hierarchical_2022} report conducting their experiments with two mesh resolutions: a coarse mesh used for the \textit{low-fidelity model} and a fine mesh used for the \textit{high-fidelity model}. The reference mesh size defined in this work corresponds to the coarse mesh. This choice is motivated by the need to keep the function evaluation time as short as possible.
\subsubsection{Variable Definition}
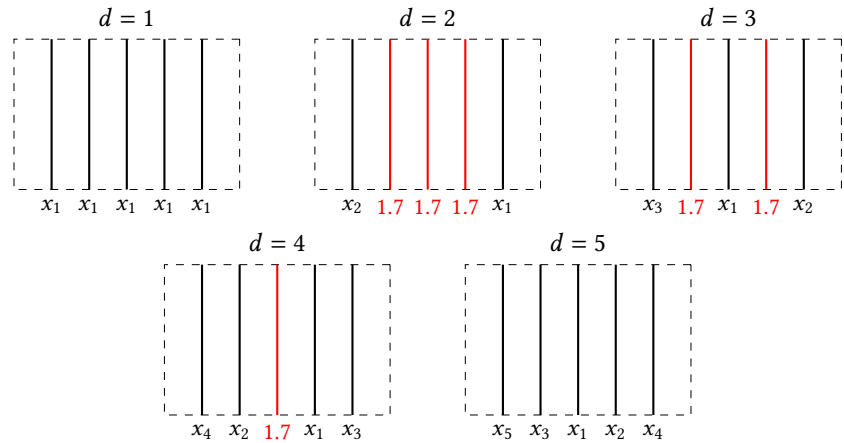
\begin{figure}[htbp]
    \centering
    \begin{tikzpicture}

\begin{scope}
	\node at (0,1.3) {$d=1$};
	\coordinate (A1) at (1.5,1);
	\coordinate (A2) at (-1.5,1);
	\coordinate (A3) at (-1.5,-1);
	\coordinate (A4) at (1.5,-1);
	\draw[ultra thin, dashed] (A1)--(A2)--(A3)--(A4)--cycle;
	
	\draw[thick] (-1,1) -- (-1,-1) node[below,font=\small] {$x_{1}$};
	\draw[thick] (-0.5,1) -- (-0.5,-1) node[below,font=\small] {$x_{1}$};
	\draw[thick] (-0,1) -- (-0,-1) node[below,font=\small] {$x_{1}$};
	\draw[thick] (0.5,1) -- (0.5,-1) node[below,font=\small] {$x_{1}$};
	\draw[thick] (1,1) -- (1,-1) node[below,font=\small] {$x_{1}$};
	
\end{scope}


\begin{scope}[xshift=4cm]
	\node at (0,1.3) {$d=2$};
	\coordinate (A1) at (1.5,1);
	\coordinate (A2) at (-1.5,1);
	\coordinate (A3) at (-1.5,-1);
	\coordinate (A4) at (1.5,-1);
	\draw[ultra thin, dashed] (A1)--(A2)--(A3)--(A4)--cycle;
	
	\draw[thick] (-1,1) -- (-1,-1) node[below,font=\small] {$x_{2}$};
	\draw[thick, red] (-0.5,1) -- (-0.5,-1) node[below,font=\small] {$1.7$};
	\draw[thick,red] (-0,1) -- (-0,-1) node[below,font=\small] {$1.7$};
	\draw[thick, red] (0.5,1) -- (0.5,-1) node[below,font=\small] {$1.7$};
	\draw[thick] (1,1) -- (1,-1) node[below,font=\small] {$x_{1}$};
	
\end{scope}

\begin{scope}[xshift=8cm]
	\node at (0,1.3) {$d=3$};
	\coordinate (A1) at (1.5,1);
	\coordinate (A2) at (-1.5,1);
	\coordinate (A3) at (-1.5,-1);
	\coordinate (A4) at (1.5,-1);
	\draw[ultra thin, dashed] (A1)--(A2)--(A3)--(A4)--cycle;
	
	\draw[thick] (-1,1) -- (-1,-1) node[below,font=\small] {$x_{3}$};
	\draw[thick, red] (-0.5,1) -- (-0.5,-1) node[below,font=\small] {$1.7$};
	\draw[thick] (-0,1) -- (-0,-1) node[below,font=\small] {$x_{1}$};
	\draw[thick, red] (0.5,1) -- (0.5,-1) node[below,font=\small] {$1.7$};
	\draw[thick] (1,1) -- (1,-1) node[below,font=\small] {$x_{2}$};
	
\end{scope}

\begin{scope}[xshift=2cm, yshift=-3cm]
	\node at (0,1.3) {$d=4$};
	\coordinate (A1) at (1.5,1);
	\coordinate (A2) at (-1.5,1);
	\coordinate (A3) at (-1.5,-1);
	\coordinate (A4) at (1.5,-1);
	\draw[ultra thin, dashed] (A1)--(A2)--(A3)--(A4)--cycle;
	
	\draw[thick] (-1,1) -- (-1,-1) node[below,font=\small] {$x_{4}$};
	\draw[thick] (-0.5,1) -- (-0.5,-1) node[below,font=\small] {$x_2$};
	\draw[thick,red] (-0,1) -- (-0,-1) node[below,font=\small] {$1.7$};
	\draw[thick] (0.5,1) -- (0.5,-1) node[below,font=\small] {$x_1$};
	\draw[thick] (1,1) -- (1,-1) node[below,font=\small] {$x_{3}$};
	
\end{scope}

\begin{scope}[xshift=6cm, yshift=-3cm]
	\node at (0,1.3) {$d=5$};
	\coordinate (A1) at (1.5,1);
	\coordinate (A2) at (-1.5,1);
	\coordinate (A3) at (-1.5,-1);
	\coordinate (A4) at (1.5,-1);
	\draw[ultra thin, dashed] (A1)--(A2)--(A3)--(A4)--cycle;
	
	\draw[thick] (-1,1) -- (-1,-1) node[below,font=\small] {$x_{5}$};
	\draw[thick] (-0.5,1) -- (-0.5,-1) node[below,font=\small] {$x_3$};
	\draw[thick] (-0,1) -- (-0,-1) node[below,font=\small] {$x_1$};
	\draw[thick] (0.5,1) -- (0.5,-1) node[below,font=\small] {$x_2$};
	\draw[thick] (1,1) -- (1,-1) node[below,font=\small] {$x_{4}$};
	
\end{scope}

\end{tikzpicture}
    \caption{Optimization problem parameterization for $d\leq5$.  Each variable is expressed in mm. In \textcolor{red}{red} are the ribs which have a predefined thickness of 1.7 mm. Each line is annotated by the design variable denoting the wall thickness of the respective rib.}
    \label{fig:three_point_bending_low_dim}
\end{figure}

Like in the previous case, we make a distinction between the \textit{lower-dimensional} and the \textit{higher-dimensional} setting because for the \textit{lower-dimensional} case, the thickness variations are coupled for several ribs and others are kept to a fixed thickness. The aforementioned description is represented in Figure \ref{fig:three_point_bending_low_dim} in the cases where $d\leq5$, the design variables account for a uniform wall thickness of the ribs.

\begin{figure}[htbp]
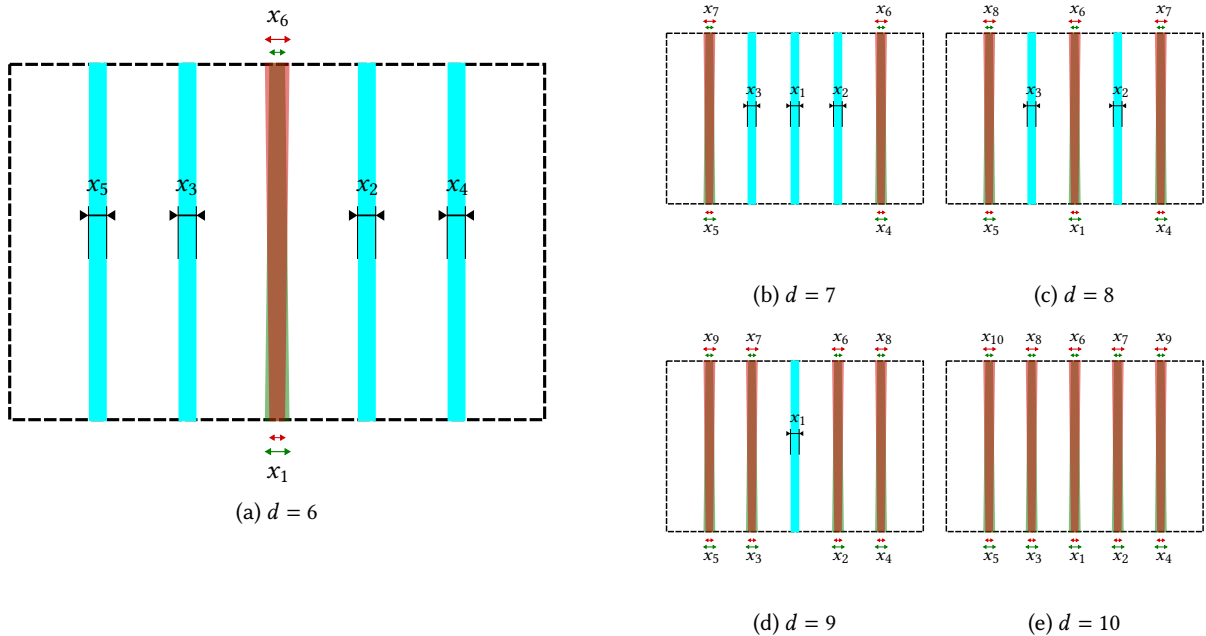


	\begin{subfigure}[t]{0.45\textwidth}
	  \vspace{0pt}
	  \includesvg[width=\linewidth]{SVGFigures/general_interpretation_sm2_2.svg}
	  \caption{$d=6$}
	  \label{fig:three_point_bending_high_dim_6}
	\end{subfigure}\hfill
	\begin{subfigure}[t]{0.45\textwidth}
	  \centering
	  \begin{subfigure}[t]{0.48\textwidth}
	    \vspace{0pt}
	    \begin{tiny}
	    	\includesvg[width=\linewidth]{SVGFigures/general_interpretation_sm2_3.svg}
	    \end{tiny}
	    \caption{$d=7$}
	    \label{fig:three_point_bending_high_dim_7}
	  \end{subfigure}\hfill
	  \begin{subfigure}[t]{0.48\textwidth}
	    \vspace{0pt}
	    \begin{tiny}
	    \includesvg[width=\linewidth]{SVGFigures/general_interpretation_sm2_4.svg}
	    \end{tiny}
	    \caption{$d=8$}
	    \label{fig:three_point_bending_high_dim_8}
	  \end{subfigure}
	
	  \vspace{0.6\baselineskip}
	
	  \begin{subfigure}[t]{0.48\textwidth}
	    \vspace{0pt}
	    \begin{tiny}
	    \includesvg[width=\linewidth]{SVGFigures/general_interpretation_sm2_5.svg}
	    \end{tiny}
	    \caption{$d=9$}
	    \label{fig:three_point_bending_high_dim_9}
	  \end{subfigure}\hfill
	  \begin{subfigure}[t]{0.48\textwidth}
	    \vspace{0pt}
	    \begin{tiny}
	    	 \includesvg[width=\linewidth]{SVGFigures/general_interpretation_sm2_6.svg}
	    \end{tiny}
	    \caption{$d=10$}
	    \label{fig:three_point_bending_high_dim_10}
	  \end{subfigure}
	\end{subfigure}

    \caption[Description of point addition to develop wall thickness profiles for each of the 5 ribs of the beam for $d=6,7,8,9,10$.]{Description of point addition used to develop wall thickness profiles for each of the five ribs of the beam for $d = 6, 7, 8, 9, 10$. For the dimensions shown, the straight rectangles colored in \textcolor{cyan}{cyan} indicate a constant thickness along the corresponding rib. In contrast, the two sets of trapezoids, colored in {\transparent{0.5}\textcolor{t1_red}{semi-transparent red}} and {\transparent{0.5}\textcolor{t1_green}{semi-transparent green}}, represent linearly increasing and decreasing thickness profiles, respectively. The superposition of these trapezoids is used to emphasize that the thickness profile of a rib may be expressed as either increasing or decreasing, depending on the variable values. For $10 < d \leq 40$, control points are added in a similar fashion as the shown sequences herein.}
\label{fig:three_point_bending_high_dim_overall}
\end{figure}

For $d > 5$, wall thickness profiles are generated for each rib by incrementally by adding control points. An example is provided in Figure \ref{fig:three_point_bending_high_dim_overall} for $6 \leq d \leq 10$. The procedure is extended to $d = 40$, such that each row of elements in the $z$-direction of a rib is assigned an independent $t_h$ value. As in the star shaped crash-box case, thickness values are interpolated with respect to the $z$-coordinate of the element barycenters, with eight rows of elements defined along this direction.

Each $x_i$ variable of this case has the same bounds, namely: $0.5 \text{ mm} \leq x_i \leq  3.0 \text{ mm}$. 

\subsubsection{Standard Mathematical Formulation}
The problem is originally formulated as a constrained optimization task:  
\begin{equation}
\begin{aligned}
\min_{\mathbf{x}} \;\; & m_{s}(\mathbf{x}) \\
\text{s.t.} \;\; & \delta(\mathbf{x}) \leq 50 \ \unit{mm}.
\end{aligned}
\label{eq:standard_problem_three_point_bending}
\end{equation}

As in the previous case, the constraint $\delta(\mathbf{x}) \leq 50 \ \unit{mm}$ imposes an intrusion limit at the point of impact. This requirement promotes an efficient mass distribution through the rib thicknesses to withstand crash loading. In crashworthiness terms, the intrusion constraint is defined to ensure that the beam neither exceeds permissible deformation levels nor fractures under the specified impact conditions, thereby satisfying both structural integrity and occupant safety requirements.

\subsubsection{Reformulation}
Similar to the star shaped crash-box benchmark, we do not employ a penalized objective function as in \citet{kaps_hierarchical_2022}. Instead, the objective was defined as a piecewise function based on feasibility, determined by the measured maximum intrusion after impact. This is expressed as:
\begin{equation}
\min_{\mathbf{x}}\text{Penalized } m_s (\mathbf{x})  =
\begin{cases}
m_s (\mathbf{x}), & \text{if } \delta (\mathbf{x}) \leq 50 \ \unit{mm}, \\
4.25952 + 10 \left( \frac{\delta}{50}-1 \right), & \text{if } \delta (\mathbf{x}) > 50 \ \unit{mm}.
\end{cases}
\label{eq:reformulated_problem_three_point_bending}
\end{equation}

\subsection{Long Crash Tube}

\subsubsection{Problem Description}
The primary objective in this study is to mitigate peak force amplification by optimizing the shape and placement of triggers in a long crash box. Modifying the trigger characteristics alters the deformation behavior of the structure, enabling better energy absorption and more controlled plastic collapse modes.

\begin{figure}[htbp]
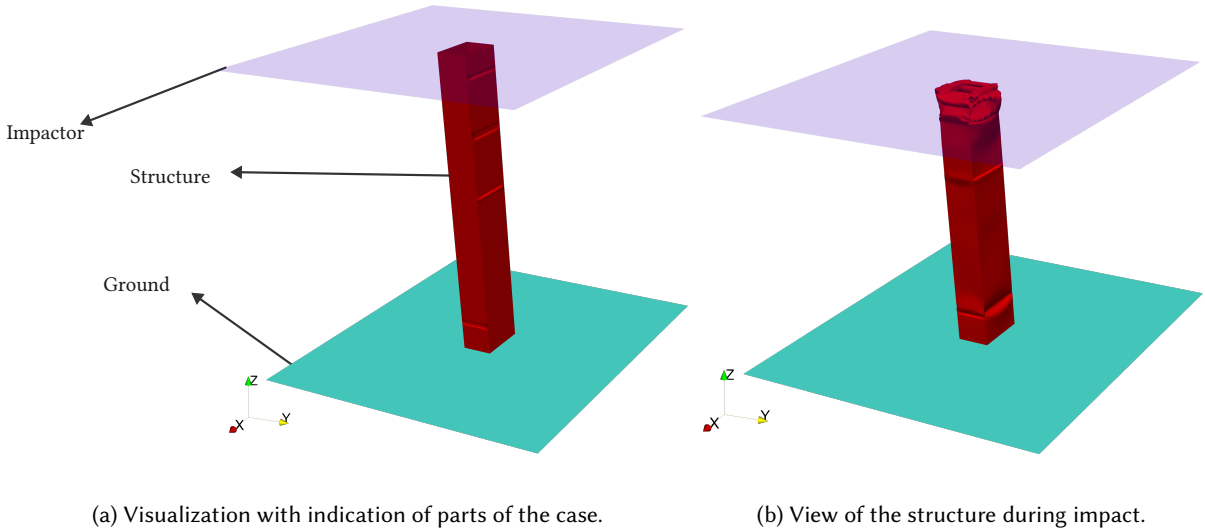

    \centering
    \begin{subfigure}[b]{0.57\linewidth}
	    \begin{footnotesize}
	    	\includesvg[width=\linewidth]{SVGFigures/crashtube_diagram.svg}
	    \end{footnotesize}
	    \caption{Visualization with indication of parts of the case.}
	    \label{fig:crashtube_representation_1}
    \end{subfigure}
    \hfill
    \begin{subfigure}[b]{0.42\linewidth}
	    	\begin{footnotesize}
	    		\includesvg[width=\linewidth]{SVGFigures/crashtube_2_diagram.svg}
	    	\end{footnotesize}
	    	\caption{View of the structure during impact.}
	    \label{fig:crashtube_representation_2}
    \end{subfigure}
    \caption{Representation of the Long Crash Tube Case.}
    \label{fig:crashtube_representation}
\end{figure}

We took inspiration from the case analyzed in \citet{kaps_hierarchical_2022}, but we add additional modifications to enable scalability. Yet the definition of the case is mostly the same. In Figure \ref{fig:crashtube_representation}, there's a representation of the objects of the case, where the impactor is represented as a solid moving towards the structure, which is clamped to the ground.
%
%


%
%
%

%
\subsubsection{Variable Definition}
\begin{figure}[htbp!]
    \centering
    \includesvg[width=0.7\linewidth]{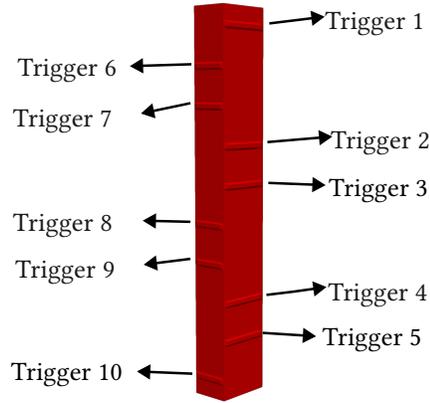}
    \caption{Labeling of triggers of the Long Crash Tube Case.}
    \label{fig:crashtube_representation_2b}
\end{figure}

To provide a clearer understanding of the design (optimization) variables involved in this study, we refer to the labeling scheme illustrated in Figure~\ref{fig:crashtube_representation_2b}. The cross-sectional geometry of the structure is rectangular, consisting of two pairs of opposing faces. We define Face A as the pair corresponding to the longer edges of the rectangle, and Face B as the pair corresponding to the shorter edges. A total of ten triggers are distributed along these faces: five on Face~A, indexed from 1 to 5, and five on Face~B, indexed from 6 to 10.

\begin{figure}[hbtp]
	\centering
    \begin{footnotesize}
    	\includesvg[width=0.62\linewidth, inkscapelatex=true]{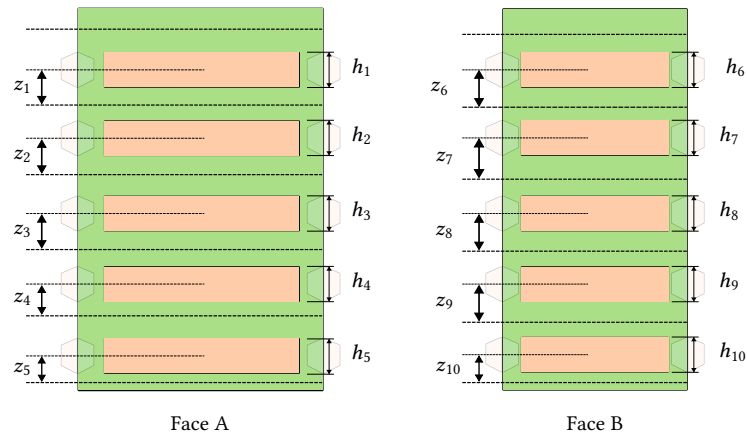}
     \end{footnotesize}
    \caption[Representation of the pairs $(z_i,h_i)$, where $z_i$ denotes the position along the $z$-axis of the barycenter of a trigger, and $h_i$ is the corresponding height.]{Representation of the pairs $(z_i,h_i)$, where $z_i$ denotes the position along the $z$-axis of the barycenter of a trigger, and $h_i$ is the corresponding height. In this representation, the triggers are shown in \textcolor{t1_peach}{light peach}, while the rest of the structure is shown in \textcolor{t1_light_green}{light green}. The $z$-position is measured from a reference point. The thick dashed lines indicate the possible limits of each $z_i$, whereas the thin dashed lines mark the $z$-position of the barycenter of a trigger. The drawing is not in scale. The triggers located at complementary faces are shown transparent.}
    \label{fig:triggers_z_h}
\end{figure}

\begin{figure}[hbtp]
	\centering
    \begin{footnotesize}
    	\includesvg[width=0.55\linewidth, inkscapelatex=true]{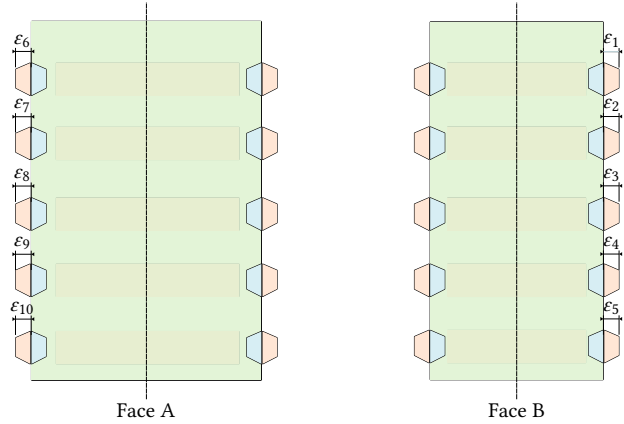}
     \end{footnotesize}
    \caption[Representation of each $\varepsilon_i$ on the structure]{Representation of each $\varepsilon_i$ on the structure. The reference point for each $\varepsilon_i$ corresponds to a plane attached to either Face A or Face B. In this figure, each $\varepsilon_i$ is measured is visualized as an outward extrusion, shown in \textcolor{pastel_peach}{pastel peach}. When $\varepsilon_i$ is negative, the measure is instead represented as an inward extrusion, shown in \textcolor{sky_blue}{sky blue}. The dashed line drawn at the middle of the structure corresponds to a symmetry axis. This means that the same $\varepsilon_i$ value is imposed to the complementary face. The drawing is not in scale. Similarly, the triggers of frontal face are shown transparent to point that from such views the triggers are present, yet not related to show the meaning of each $\varepsilon_i$.}
    \label{fig:triggers_varepsilon}
\end{figure}

Each trigger is characterized by a triplet \((z_i, \varepsilon_i, h_i)\), where \(z_i\) denotes the axial (along the \(z\)-axis) position of the trigger’s centroid relative to a fixed reference point, \(\varepsilon_i\) represents the trigger's protrusion (depth) from the face surface, and \(h_i\) corresponds to its vertical extent (height). The representation of the variables on the structure are shown in Figures \ref{fig:triggers_z_h} and \ref{fig:triggers_varepsilon}. These parameters serve as the primary design variables in the optimization process. The bounds of the triplet are the following:
\begin{align*}
    -40 \text{ mm} \leq z_i &\leq 40 \text{ mm}, \\
    -4 \text{ mm} \leq \varepsilon_i &\leq 4 \text{ mm}, \\
    0 \text{ mm} \leq h_i &\leq 16 \text{ mm}.
\end{align*}
%
%
Finally, we developed a table of equivalences, wherein it is stipulated which design variable $x_{i}$ controls the triplets $(z_i, \varepsilon_i, h_i)$ for different dimensions. We split such a table of equivalence into two parts, which are shown in Tables \ref{tab:equiv_table_1} and \ref{tab:equiv_table_2}. This splitting is worthy since for $d\leq15$, the design variables control the triggers of the two faces in a mirrored way, whereas for $15<d\leq 30$, the triggers of both faces are controlled independently.


\subsubsection{Standard Mathematical Formulation}
This problem is treated as unconstrained and is formulated as:  

\begin{equation}
\min_{\mathbf{x}} \;\; \text{LU}(\mathbf{x}) = \left | \frac{F_{\text{peak}}}{F_{\text{mean}}} \right |,
\label{eq:standard_problem_crash_tube}
\end{equation}
\nomenclature{LU}{Load Uniformity}{kN/kN}{}
\nomenclature{$F_{\text{peak}}$}{Peak Force during impact}{kN}{}
\nomenclature{$F_{\text{mean}}$}{Mean Force over the loading duration}{kN}{}
where $\text{LU}$ denotes the load uniformity, $F_{\text{peak}}$ is the maximum force recorded during impact, and $F_{\text{mean}}$ is the mean force over the loading duration.  

Minimizing load uniformity in this context reduces the amplification of peak forces relative to the average crash load. High peak forces are undesirable, as they can cause sudden stress concentrations and localized failures, whereas smoother force distributions enable the structure to dissipate energy more steadily throughout the deformation process. 
\section{Conclusions and future work}
We presented a benchmarking suite for mechanical design and crashworthiness optimization, designed to foster reproducibility, comparability, and scalability in black-box optimization applied to engineering simulations. The benchmarks are modular and moderately lightweight, supporting seamless integration with external optimizers through standardized interfaces, thereby minimizing setup overhead.

At present, the suite focuses on single-objective problems involving shell-based finite element models. These cover a relevant subset of shape optimization tasks but do not yet capture the full complexity of many practical design scenarios. To move towards a more general and representative evaluation platform, we plan to extend the framework along two axes.

First, we will implement problem instances by applying transformations in a manner similar to the noiseless black-box optimization benchmarks (\texttt{BBOB}) described in \cite{bbob2019}. We believe this addition will make the suite particularly valuable to the community, as other real-world-inspired benchmarking suites rarely account for such variations. This will enable a more robust and fair evaluation of algorithms, since no universal shortcut will exist for tailoring an optimizer to a single fixed problem.

Second, we acknowledge that a major limitation of the current setup is the high simulation time displayed in Table \ref{tab:runtime}, which can restrict the scale of experimental evaluations. To address this, we plan to incorporate surrogate models and aliasing strategies into the benchmarking repository, enabling faster and more extensive testing. In addition, we will embed these surrogates within multi-fidelity evaluations to reflect different levels of model precision and better approximate real engineering design processes.

Finally, beyond the optimization interface itself, we will broaden the benchmark set by integrating more representative components and assemblies, and by incorporating datasets from public initiatives and industrial collaborations.

\begin{acks}
We thank the Chair of Computational Solid Mechanics (CSM) at the Technical University of Munich, and in particular Arne Kaps and Paolo Ascia, for their many insightful discussions. We also thank Fei Fan Li for her contributions to the initial development of the benchmarks.
\end{acks}

\bibliographystyle{ACM-Reference-Format} 
\bibliography{refs} 

@article{fang_design_2017,
	title = {On design optimization for structural crashworthiness and its state of the art},
	volume = {55},
	issn = {1615-147X, 1615-1488},
	url = {http://link.springer.com/10.1007/s00158-016-1579-y},
	doi = {10.1007/s00158-016-1579-y},
	language = {en},
	number = {3},
	urldate = {2025-06-13},
	journal = {Structural and Multidisciplinary Optimization},
	author = {Fang, Jianguang and Sun, Guangyong and Qiu, Na and Kim, Nam H. and Li, Qing},
	month = mar,
	year = {2017},
	pages = {1091--1119},
}

@article{zabaras_continuum_2003,
	title = {A continuum sensitivity method for the design of multi-stage metal forming processes},
	volume = {45},
	copyright = {https://www.elsevier.com/tdm/userlicense/1.0/},
	issn = {00207403},
	url = {https://linkinghub.elsevier.com/retrieve/pii/S0020740303000481},
	doi = {10.1016/S0020-7403(03)00048-1},
	language = {en},
	number = {2},
	urldate = {2025-06-13},
	journal = {International Journal of Mechanical Sciences},
	author = {Zabaras, Nicholas and Ganapathysubramanian, Shankar and Li, Qing},
	month = feb,
	year = {2003},
	pages = {325--358},
}

@book{dubois_vehicle_2004,
	address = {Southfield, Michigan, United States of America},
	title = {Vehicle {Crashworthiness} and {Occupant} {Protection}},
	language = {en},
	publisher = {American Iron and Steel Institute},
	author = {DuBois, Paul and Chou, Clifford and Fileta, Bahig and Khalil, Tawfik and King, Albert and Mahmood, Hikmat and Mertz, Harold and Wismans, Jac and Prasad, Priya and Belwafa, Jamel},
	year = {2004},
}

@article{duddeck_multidisciplinary_2008,
	title = {Multidisciplinary optimization of car bodies},
	volume = {35},
	copyright = {http://www.springer.com/tdm},
	issn = {1615-147X, 1615-1488},
	url = {http://link.springer.com/10.1007/s00158-007-0130-6},
	doi = {10.1007/s00158-007-0130-6},
	language = {en},
	number = {4},
	urldate = {2025-06-13},
	journal = {Structural and Multidisciplinary Optimization},
	author = {Duddeck, Fabian},
	month = apr,
	year = {2008},
	pages = {375--389},
}

@article{kudela_critical_2022,
	title = {A critical problem in benchmarking and analysis of evolutionary computation methods},
	volume = {4},
	issn = {2522-5839},
	url = {https://www.nature.com/articles/s42256-022-00579-0},
	doi = {10.1038/s42256-022-00579-0},
	language = {en},
	number = {12},
	urldate = {2025-04-21},
	journal = {Nature Machine Intelligence},
	author = {Kudela, Jakub},
	month = dec,
	year = {2022},
	pages = {1238--1245},
}

@article{hansen_coco_2021,
	title = {{COCO}: {A} {Platform} for {Comparing} {Continuous} {Optimizers} in a {Black}-{Box} {Setting}},
	volume = {36},
	issn = {1055-6788, 1029-4937},
	shorttitle = {{COCO}},
	url = {http://arxiv.org/abs/1603.08785},
	doi = {10.1080/10556788.2020.1808977},
	number = {1},
	urldate = {2025-06-13},
	journal = {Optimization Methods and Software},
	author = {Hansen, Nikolaus and Auger, Anne and Ros, Raymond and Mersmann, Olaf and Tušar, Tea and Brockhoff, Dimo},
	month = jan,
	year = {2021},
	note = {arXiv:1603.08785 [cs]},
	pages = {114--144},
}

@article{CEC_Michalewicz2000,
author = {Michalewicz, Z. and Deb, K. and Schmidt, M. and Stidsen, T.},
title = {Test-case generator for nonlinear continuous parameter optimization techniques},
year = {2000},
issue_date = {September 2000},
publisher = {IEEE Press},
volume = {4},
number = {3},
issn = {1089-778X},
url = {https://doi.org/10.1109/4235.873232},
doi = {10.1109/4235.873232},
journal = {Trans. Evol. Comp},
month = sep,
pages = {197–215},
numpages = {19}
}

@article{bliek_benchmarking_2023,
	title = {Benchmarking surrogate-based optimisation algorithms on expensive black-box functions},
	volume = {147},
	issn = {15684946},
	url = {https://linkinghub.elsevier.com/retrieve/pii/S1568494623007627},
	doi = {10.1016/j.asoc.2023.110744},
	language = {en},
	urldate = {2025-04-29},
	journal = {Applied Soft Computing},
	author = {Bliek, Laurens and Guijt, Arthur and Karlsson, Rickard and Verwer, Sicco and De Weerdt, Mathijs},
	month = nov,
	year = {2023},
	pages = {110744},
}

@article{gorissen_surrogate_2010,
	title = {A {Surrogate} {Modeling} and {Adaptive} {Sampling} {Toolbox} for {Computer} {Based} {Design}},
	volume = {11},
	journal = {J. Mach. Learn. Res.},
	author = {Gorissen, Dirk and Couckuyt, Ivo and Demeester, Piet and Dhaene, Tom and Crombecq, Karel},
	year = {2010},
	note = {ISBN: 1532-4435
Publisher: JMLR.org},
	pages = {2051--2055},
}

@incollection{daniels_suite_2018,
	address = {Cham},
	title = {A {Suite} of {Computationally} {Expensive} {Shape} {Optimisation} {Problems} {Using} {Computational} {Fluid} {Dynamics}},
	volume = {11102},
	isbn = {978-3-319-99258-7 978-3-319-99259-4},
	url = {http://link.springer.com/10.1007/978-3-319-99259-4_24},
	language = {en},
	urldate = {2025-06-13},
	booktitle = {Parallel {Problem} {Solving} from {Nature} – {PPSN} {XV}},
	publisher = {Springer International Publishing},
	author = {Daniels, Steven J. and Rahat, Alma A. M. and Everson, Richard M. and Tabor, Gavin R. and Fieldsend, Jonathan E.},
	editor = {Auger, Anne and Fonseca, Carlos M. and Lourenço, Nuno and Machado, Penousal and Paquete, Luís and Whitley, Darrell},
	year = {2018},
	doi = {10.1007/978-3-319-99259-4_24},
	note = {Series Title: Lecture Notes in Computer Science},
	pages = {296--307},
}

@techreport{jones2008large,
  author       = {Jones, Donald R.},
  title        = {Large-Scale Multi-Disciplinary Mass Optimization in the Auto Industry},
  institution  = {MOPTA 2008},
  address      = {Detroit, MI, USA},
  year         = {2008}
}

@incollection{raponi_global_2025,
	title = {Global {Sensitivity} {Analysis} {Is} {Not} {Always} {Beneficial} for {Evolutionary} {Computation}: {A} {Study} in {Engineering} {Design}},
	isbn = {978-981-96-2540-6},
	url = {https://doi.org/10.1007/978-981-96-2540-6_2},
	booktitle = {Explainable {AI} for {Evolutionary} {Computation}},
	publisher = {Springer Nature Singapore},
	author = {Raponi, Elena and Olarte Rodriguez, Ivan and van Stein, Niki},
	editor = {van Stein, Niki and Kononova, Anna V.},
	year = {2025},
	doi = {10.1007/978-981-96-2540-6_2},
	pages = {13--40},
}

@article{hunkeler_shape_2013,
	title = {Shape optimisation for crashworthiness followed by a robustness analysis with respect to shape variables: {Example} of a front rail},
	volume = {48},
	copyright = {http://www.springer.com/tdm},
	issn = {1615-147X, 1615-1488},
	shorttitle = {Shape optimisation for crashworthiness followed by a robustness analysis with respect to shape variables},
	url = {http://link.springer.com/10.1007/s00158-013-0903-z},
	doi = {10.1007/s00158-013-0903-z},
	language = {en},
	number = {2},
	urldate = {2024-08-28},
	journal = {Structural and Multidisciplinary Optimization},
	author = {Hunkeler, Stephan and Duddeck, Fabian and Rayamajhi, Milan and Zimmer, Hans},
	month = aug,
	year = {2013},
	pages = {367--378},
}

@article{kaps_hierarchical_2022,
	title = {A hierarchical kriging approach for multi-fidelity optimization of automotive crashworthiness problems},
	volume = {65},
	issn = {1615-147X, 1615-1488},
	url = {https://link.springer.com/10.1007/s00158-022-03211-2},
	doi = {10.1007/s00158-022-03211-2},
	language = {en},
	number = {4},
	urldate = {2024-08-28},
	journal = {Structural and Multidisciplinary Optimization},
	author = {Kaps, Arne and Czech, Catharina and Duddeck, Fabian},
	month = apr,
	year = {2022},
	pages = {114},
}

@inproceedings{Volz2007,
  author       = {Volz, Karl-Heinz and Frodl, Bernhard and Dirschmid, Ferdinand and Stryczek, Rafael and Zimmer, Hans},
  title        = {Optimizing topology and shape for crashworthiness in vehicle product development},
  booktitle    = {International Automotive Body Congress (IABC)},
  year         = {2007},
  month        = jun,
  day          = {17--19},
  address      = {Berlin, Germany},
}

@article{geuzaine_gmsh_2009,
	title = {Gmsh: {A} 3‐{D} finite element mesh generator with built‐in pre‐ and post‐processing facilities},
	volume = {79},
	copyright = {http://onlinelibrary.wiley.com/termsAndConditions\#vor},
	issn = {0029-5981, 1097-0207},
	shorttitle = {Gmsh},
	url = {https://onlinelibrary.wiley.com/doi/10.1002/nme.2579},
	doi = {10.1002/nme.2579},
	language = {en},
	number = {11},
	urldate = {2025-09-08},
	journal = {International Journal for Numerical Methods in Engineering},
	author = {Geuzaine, Christophe and Remacle, Jean‐François},
	month = sep,
	year = {2009},
	pages = {1309--1331},
}

@Article{         harris2020array,
 title         = {Array programming with {NumPy}},
 author        = {Charles R. Harris and K. Jarrod Millman and St{\'{e}}fan J.
                 van der Walt and Ralf Gommers and Pauli Virtanen and David
                 Cournapeau and Eric Wieser and Julian Taylor and Sebastian
                 Berg and Nathaniel J. Smith and Robert Kern and Matti Picus
                 and Stephan Hoyer and Marten H. van Kerkwijk and Matthew
                 Brett and Allan Haldane and Jaime Fern{\'{a}}ndez del
                 R{\'{i}}o and Mark Wiebe and Pearu Peterson and Pierre
                 G{\'{e}}rard-Marchant and Kevin Sheppard and Tyler Reddy and
                 Warren Weckesser and Hameer Abbasi and Christoph Gohlke and
                 Travis E. Oliphant},
 year          = {2020},
 month         = sep,
 journal       = {Nature},
 volume        = {585},
 number        = {7825},
 pages         = {357--362},
 doi           = {10.1038/s41586-020-2649-2},
 publisher     = {Springer Science and Business Media {LLC}},
 url           = {https://doi.org/10.1038/s41586-020-2649-2}
}

@article{jain_evolutionary_2014,
	title = {An {Evolutionary} {Many}-{Objective} {Optimization} {Algorithm} {Using} {Reference}-{Point} {Based} {Nondominated} {Sorting} {Approach}, {Part} {II}: {Handling} {Constraints} and {Extending} to an {Adaptive} {Approach}},
	volume = {18},
	issn = {1941-0026},
	shorttitle = {An {Evolutionary} {Many}-{Objective} {Optimization} {Algorithm} {Using} {Reference}-{Point} {Based} {Nondominated} {Sorting} {Approach}, {Part} {II}},
	url = {https://ieeexplore.ieee.org/document/6595567/},
	doi = {10.1109/TEVC.2013.2281534},
	number = {4},
	urldate = {2025-09-25},
	journal = {IEEE Transactions on Evolutionary Computation},
	author = {Jain, Himanshu and Deb, Kalyanmoy},
	month = aug,
	year = {2014},
	pages = {602--622},
}

@article{tanabe_easy_to_use_2020,
	title = {An easy-to-use real-world multi-objective optimization problem suite},
	volume = {89},
	issn = {1568-4946},
	url = {https://www.sciencedirect.com/science/article/pii/S1568494620300181},
	doi = {10.1016/j.asoc.2020.106078},
	urldate = {2025-09-25},
	journal = {Applied Soft Computing},
	author = {Tanabe, Ryoji and Ishibuchi, Hisao},
	month = apr,
	year = {2020},
	pages = {106078},
}

@inproceedings{kohira_proposal_2018,
	address = {Kyoto Japan},
	title = {Proposal of benchmark problem based on real-world car structure design optimization},
	isbn = {978-1-4503-5764-7},
	url = {https://dl.acm.org/doi/10.1145/3205651.3205702},
	doi = {10.1145/3205651.3205702},
	language = {en},
	urldate = {2025-09-23},
	booktitle = {Proceedings of the {Genetic} and {Evolutionary} {Computation} {Conference} {Companion}},
	publisher = {ACM},
	author = {Kohira, Takehisa and Kemmotsu, Hiromasa and Akira, Oyama and Tatsukawa, Tomoaki},
	month = jul,
	year = {2018},
	pages = {183--184},
}

@article{he_dafoam_2020,
	title = {{DAFoam}: {An} {Open}-{Source} {Adjoint} {Framework} for {Multidisciplinary} {Design} {Optimization} with {OpenFOAM}},
	volume = {58},
	issn = {0001-1452, 1533-385X},
	shorttitle = {{DAFoam}},
	url = {https://arc.aiaa.org/doi/10.2514/1.J058853},
	doi = {10.2514/1.J058853},
	language = {en},
	number = {3},
	urldate = {2025-09-25},
	journal = {AIAA Journal},
	author = {He, Ping and Mader, Charles A. and Martins, Joaquim R. R. A. and Maki, Kevin J.},
	month = mar,
	year = {2020},
	pages = {1304--1319},
}

@article{he_aerodynamic_2018,
	title = {An aerodynamic design optimization framework using a discrete adjoint approach with {OpenFOAM}},
	volume = {168},
	issn = {00457930},
	url = {https://linkinghub.elsevier.com/retrieve/pii/S0045793018302020},
	doi = {10.1016/j.compfluid.2018.04.012},
	language = {en},
	urldate = {2025-09-25},
	journal = {Computers \& Fluids},
	author = {He, Ping and Mader, Charles A. and Martins, Joaquim R.R.A. and Maki, Kevin J.},
	month = may,
	year = {2018},
	pages = {285--303},
}

@article{IOHexperimenter,
  author = {Jacob de Nobel and
               Furong Ye and
               Diederick Vermetten and
               Hao Wang and
               Carola Doerr and
               Thomas B{\"{a}}ck},
  title = {{IOHexperimenter: Benchmarking Platform for Iterative Optimization Heuristics}},
  journal = {arXiv e-prints:2111.04077},
  archivePrefix = "arXiv",
  eprint = {2111.04077},
  year = 2021,
  month = Nov,
  url = {https://arxiv.org/abs/2111.04077}
}

@article{rios_derivative-free_2013,
	title = {Derivative-free optimization: a review of algorithms and comparison of software implementations},
	volume = {56},
	issn = {0925-5001, 1573-2916},
	shorttitle = {Derivative-free optimization},
	url = {https://link.springer.com/10.1007/s10898-012-9951-y},
	doi = {10.1007/s10898-012-9951-y},
	language = {en},
	number = {3},
	urldate = {2025-11-12},
	journal = {Journal of Global Optimization},
	author = {Rios, Luis Miguel and Sahinidis, Nikolaos V.},
	month = jul,
	year = {2013},
	pages = {1247--1293},
}

@book{audet_derivative-free_2017,
	address = {Cham},
	series = {Springer {Series} in {Operations} {Research} and {Financial} {Engineering}},
	title = {Derivative-{Free} and {Blackbox} {Optimization}},
	copyright = {http://www.springer.com/tdm},
	isbn = {978-3-319-68912-8 978-3-319-68913-5},
	url = {http://link.springer.com/10.1007/978-3-319-68913-5},
	urldate = {2025-11-12},
	publisher = {Springer International Publishing},
	author = {Audet, Charles and Hare, Warren},
	year = {2017},
	doi = {10.1007/978-3-319-68913-5},
}

@article{wolpert_no_1997,
	title = {No free lunch theorems for optimization},
	volume = {1},
	issn = {1941-0026},
	url = {https://ieeexplore.ieee.org/document/585893},
	doi = {10.1109/4235.585893},
	number = {1},
	urldate = {2025-11-12},
	journal = {IEEE Transactions on Evolutionary Computation},
	author = {Wolpert, D.H. and Macready, W.G.},
	month = apr,
	year = {1997},
	pages = {67--82},
}

@misc{doerr_iohprofiler_2018,
	title = {{IOHprofiler}: {A} {Benchmarking} and {Profiling} {Tool} for {Iterative} {Optimization} {Heuristics}},
	shorttitle = {{IOHprofiler}},
	url = {http://arxiv.org/abs/1810.05281},
	doi = {10.48550/arXiv.1810.05281},
	urldate = {2025-11-12},
	publisher = {arXiv},
	author = {Doerr, Carola and Wang, Hao and Ye, Furong and Rijn, Sander van and Bäck, Thomas},
	month = oct,
	year = {2018},
	note = {arXiv:1810.05281 [cs]},
}

@techreport{bbob2019,
    author = {Finck, Steffen and Hansen, Nikolaus and Ros, Raymond and Auger, Anne},
    title = {Real-Parameter Black-Box Optimization Benchmarking 2009: Noiseless Functions Definitions},
    institution = {INRIA},
    year = {2009},
    number = {RR-6829},
    note = {Updated version as of February 2019},
    url = {https://inria.hal.science/inria-00362633v2/document}
}

@software{openradioss2025,
  title = {OpenRadioss},
  author = {{Altair and Contributors}},
  year = {2025},
  note = {GitHub commit 6c16e4b},
  url = {https://github.com/OpenRadioss/OpenRadioss}
}

\appendix
\section{Material Properties}
In Table \ref{tab:Material_properties_1} and \ref{tab:Material_properties_2}, the material properties for each of the problems is shown. We tried to use the same formulations from the respective sources of the problems and adapted the original formulations into OpenRadioss framework.
\begin{table}[h!]
\centering
\caption{Material properties and strain rate parameters for Problems 1 and 3 (Star Box and Long Crash Tube). The same parameters are listed in \citet{kaps_hierarchical_2022}.}
\label{tab:Material_properties_1}
\begin{tabular}{|l|c|c|}
\hline
\textbf{Parameter} & \textbf{Symbol} & \textbf{Value} \\
\hline
\nomenclature{$E$}{Young's Modulus}{GPa}{} Young’s modulus & $E$ & 200 GPa \\
\nomenclature{$\nu$}{Poisson's ratio}{}{}Poisson’s ratio & $\nu$ & 0.3 \\
\nomenclature{$\rho$}{Mass density}{\unit{kg/m^3}}{}Mass density & $\rho$ & $7830 \,\frac{\text{kg}}{\text{m}^3}$ \\
\nomenclature{$\sigma_y$}{Yield Strength}{MPa}{}Yield strength & $\sigma_y$ & 360 MPa \\
Strainrate model & & Cowper-Symmonds \\
Strainrate parameter $C$ & $C$ & 40 \\
Strainrate parameter $p$ & $p$ & 5 \\
Plasticity & & Linear Piecewise (see Figure \ref{fig:plasticity_steel}) \\
\hline
\end{tabular}

\end{table}

\begin{table}[h!]
\centering
\caption{Material properties and strain rate parameters for problem 2 (Three Point Bending of Layered Beam). The same parameters are listed in \citet{kaps_hierarchical_2022}.}
\label{tab:Material_properties_2}
\begin{tabular}{|l|c|c|}
\hline
\textbf{Parameter} & \textbf{Symbol} & \textbf{Value} \\
\hline
Young’s modulus & $E$ & 70 GPa \\
Poisson’s ratio & $\nu$ & 0.33 \\
Mass density & $\rho$ & $2700 \,\frac{\text{kg}}{\text{m}^3}$ \\
Yield strength & $\sigma_y$ & 180 MPa \\
Plasticity & & Linear Piecewise (see Figure \ref{fig:plasticity_aluminum})\\
\hline
\end{tabular}

\end{table}
\begin{figure}[htpb]
    \centering
    \begin{subfigure}[b]{0.45\linewidth}
    \centering
\begin{tikzpicture}
\begin{axis}[
    width=\linewidth,
    height=7cm,
    xlabel={Effective Plastic Strain [adim]},
    ylabel={Effective Stress [MPa]},
    grid=both,
    grid style={gray!30},
    thick,
    mark size=2pt,
    every axis plot/.append style={very thick},
    xtick={0,0.025,0.05,0.075,0.1,0.125,0.15,0.175},
    tick label style={/pgf/number format/fixed},
    xmin=0, xmax=0.190,
    ymin=360, ymax=590,
]
\addplot+[
    mark=*,
] coordinates {
    (0.000, 366)
    (0.025, 424)
    (0.049, 476)
    (0.072, 507)
    (0.095, 529)
    (0.118, 546)
    (0.140, 559)
    (0.182, 584)
};
\end{axis}
\end{tikzpicture}
    \caption{Plasticity Curve Steel}
    \label{fig:plasticity_steel}
    \end{subfigure}
    \hfill
    \begin{subfigure}[b]{0.45\linewidth}
    \centering
\begin{tikzpicture}
\begin{axis}[
    width=\linewidth,
    height=7cm,
    xlabel={Effective Plastic Strain [adim]},
    ylabel={Effective Stress [MPa]},
    grid=both,
    grid style={gray!30},
    thick,
    mark size=2pt,
    every axis plot/.append style={very thick},
    xmin=0, xmax=0.5,
    ymin=170, ymax=260,
]
\addplot+[
    mark=*,
] coordinates {
    (0, 180)
    (0.01, 190)
    (0.02, 197)
    (0.05, 211.5)
    (0.1, 225.8)
    (0.15, 233.6)
    (0.2, 238.5)
    (0.4, 248.5)
};
\end{axis}
\end{tikzpicture}
    \caption{Plasticity Curve Aluminum}
    \label{fig:plasticity_aluminum}
    \end{subfigure}
    \caption{Plasticity Curves used for the Material Models. Adapted from \citet{kaps_hierarchical_2022}.}
    \label{fig:plasticity}
\end{figure}
\section{Simulation Setup}
In Table \ref{tab:sim-settings} we show the settings of the simulations for the impactor, the volume of the structure as well as the simulation time. As shown there, we kept the same shell element formulation and the same contact formulation (the surface to surface contact) for all the problems. In problem 1, since the volume of the structure changes, we show a range instead.
\begin{table}[htpb!]
\centering
\caption{Settings for the simulation for each of the problems.}
\begin{tabular}{|l|lll|}
\hline
\textbf{Parameter}  & \multicolumn{1}{l|}{\textbf{Problem 1}} & \multicolumn{1}{l|}{\textbf{Problem 2}} & \textbf{Problem 3} \\ \hline
Element Formulation      & \multicolumn{3}{l|}{Belytschko-Lin-Tsay}                        \\ \hline
Contact Formulation & \multicolumn{3}{l|}{\texttt{/INTER/TYPE24} from Radioss}                              \\ \hline
Height [mm]          & \multicolumn{1}{l|}{120}    & \multicolumn{1}{l|}{120} & 800   \\ \hline
Width {[}mm{]}           & \multicolumn{1}{l|}{60-120} & \multicolumn{1}{l|}{800} & 120   \\ \hline
Depth {[}mm{]}           & \multicolumn{1}{l|}{60-120} & \multicolumn{1}{l|}{80}  & 80    \\ \hline
$v_{I,0}$ {[}km/hr{]}    & \multicolumn{1}{l|}{25.2}   & \multicolumn{1}{l|}{36}  & 30 \\ \hline
\nomenclature{$m_{I}$}{Mass of the impactor}{kg}{}$m_{I}$ {[}kg{]}         & \multicolumn{1}{l|}{250}    & \multicolumn{1}{l|}{86}  & 300   \\ \hline
Impactor Radius {[}mm{]} & \multicolumn{1}{l|}{}       & \multicolumn{1}{l|}{36}  &       \\ \hline
Simulation Time {[}ms{]} & \multicolumn{1}{l|}{45}     & \multicolumn{1}{l|}{40}  & 45    \\ \hline
\end{tabular}%
\label{tab:sim-settings}
\end{table}

\section{Equivalence Matrices to define variables for the Long Crash Tube case}
\label{sec:Tables of Equivalences}
The equivalences are shown in the tables \ref{tab:equiv_table_1} and \ref{tab:equiv_table_2}. 
%
\begin{longtable}[c]{|l|l|l|l|l|l|l|l|l|l|l|l|l|l|l|l|}
\hline
Variable/$d$ &
  1 &
  2 &
  3 &
  4 &
  5 &
  6 &
  7 &
  8 &
  9 &
  10 &
  11 &
  12 &
  13 &
  14 &
  15 \\ \hline \hline
\endfirsthead
\endhead
$z_{1}$ &
  $x_{1}$ &
  $x_{1}$ &
  $x_{1}$ &
  $x_{1}$ &
  $x_{1}$ &
  $x_{1}$ &
  $x_{1}$ &
  $x_{1}$ &
  $x_{1}$ &
  $x_{1}$ &
  $x_{1}$ &
  $x_{1}$ &
  $x_{1}$ &
  $x_{1}$ &
  $x_{1}$ \\ \hline
$\varepsilon_{1}$ &
  0 &
  $x_{2}$ &
  $x_{2}$ &
  $x_{2}$ &
  $x_{2}$ &
  $x_{2}$ &
  $x_{2}$ &
  $x_{2}$ &
  $x_{2}$ &
  $x_{2}$ &
  $x_{2}$ &
  $x_{2}$ &
  $x_{2}$ &
  $x_{2}$ &
  $x_{2}$ \\ \hline
$h_{1}$ &
  8 &
  0 &
  $x_{3}$ &
  $x_{3}$ &
  $x_{3}$ &
  $x_{3}$ &
  $x_{3}$ &
  $x_{3}$ &
  $x_{3}$ &
  $x_{3}$ &
  $x_{3}$ &
  $x_{3}$ &
  $x_{3}$ &
  $x_{3}$ &
  $x_{3}$ \\ \hline
$z_{2}$ &
  0 &
  0 &
  0 &
  $x_{4}$ &
  $x_{4}$ &
  $x_{4}$ &
  $x_{4}$ &
  $x_{4}$ &
  $x_{4}$ &
  $x_{4}$ &
  $x_{4}$ &
  $x_{4}$ &
  $x_{4}$ &
  $x_{4}$ &
  $x_{4}$ \\ \hline
$\varepsilon_{2}$ &
  0 &
  0 &
  0 &
  0 &
  $x_{5}$ &
  $x_{5}$ &
  $x_{5}$ &
  $x_{5}$ &
  $x_{5}$ &
  $x_{5}$ &
  $x_{5}$ &
  $x_{5}$ &
  $x_{5}$ &
  $x_{5}$ &
  $x_{5}$ \\ \hline
$h_{2}$ &
  8 &
  8 &
  8 &
  8 &
  8 &
  $x_{6}$ &
  $x_{6}$ &
  $x_{6}$ &
  $x_{6}$ &
  $x_{6}$ &
  $x_{6}$ &
  $x_{6}$ &
  $x_{6}$ &
  $x_{6}$ &
  $x_{6}$ \\ \hline
$z_{3}$ &
  0 &
  0 &
  0 &
  0 &
  0 &
  0 &
  $x_{7}$ &
  $x_{7}$ &
  $x_{7}$ &
  $x_{7}$ &
  $x_{7}$ &
  $x_{7}$ &
  $x_{7}$ &
  $x_{7}$ &
  $x_{7}$ \\ \hline
$\varepsilon_{3}$ &
  0 &
  0 &
  0 &
  0 &
  0 &
  0 &
  0 &
  $x_{8}$ &
  $x_{8}$ &
  $x_{8}$ &
  $x_{8}$ &
  $x_{8}$ &
  $x_{8}$ &
  $x_{8}$ &
  $x_{8}$ \\ \hline
$h_{3}$ &
  8 &
  8 &
  8 &
  8 &
  8 &
  8 &
  8 &
  8 &
  $x_{9}$ &
  $x_{9}$ &
  $x_{9}$ &
  $x_{9}$ &
  $x_{9}$ &
  $x_{9}$ &
  $x_{9}$ \\ \hline
$z_{4}$ &
  0 &
  0 &
  0 &
  0 &
  0 &
  0 &
  0 &
  0 &
  0 &
  $x_{10}$ &
  $x_{10}$ &
  $x_{10}$ &
  $x_{10}$ &
  $x_{10}$ &
  $x_{10}$ \\ \hline
$\varepsilon_{4}$ &
  0 &
  0 &
  0 &
  0 &
  0 &
  0 &
  0 &
  0 &
  0 &
  0 &
  $x_{11}$ &
  $x_{11}$ &
  $x_{11}$ &
  $x_{11}$ &
  $x_{11}$ \\ \hline
$h_{4}$ &
  8 &
  8 &
  8 &
  8 &
  8 &
  8 &
  8 &
  8 &
  8 &
  8 &
  8 &
  $x_{12}$ &
  $x_{12}$ &
  $x_{12}$ &
  $x_{12}$ \\ \hline
$z_{5}$ &
  0 &
  0 &
  0 &
  0 &
  0 &
  0 &
  0 &
  0 &
  0 &
  0 &
  0 &
  0 &
  $x_{13}$ &
  $x_{13}$ &
  $x_{13}$ \\ \hline
$\varepsilon_{5}$ &
  0 &
  0 &
  0 &
  0 &
  0 &
  0 &
  0 &
  0 &
  0 &
  0 &
  0 &
  0 &
  0 &
  $x_{14}$ &
  $x_{14}$ \\ \hline
$h_{5}$ &
  8 &
  8 &
  8 &
  8 &
  8 &
  8 &
  8 &
  8 &
  8 &
  8 &
  8 &
  8 &
  8 &
  8 &
  $x_{15}$ \\ \hline
$z_{6}$ &
  $x_{1}$ &
  $x_{1}$ &
  $x_{1}$ &
  $x_{1}$ &
  $x_{1}$ &
  $x_{1}$ &
  $x_{1}$ &
  $x_{1}$ &
  $x_{1}$ &
  $x_{1}$ &
  $x_{1}$ &
  $x_{1}$ &
  $x_{1}$ &
  $x_{1}$ &
  $x_{1}$ \\ \hline
$\varepsilon_{6}$ &
  0 &
  $x_{2}$ &
  $x_{2}$ &
  $x_{2}$ &
  $x_{2}$ &
  $x_{2}$ &
  $x_{2}$ &
  $x_{2}$ &
  $x_{2}$ &
  $x_{2}$ &
  $x_{2}$ &
  $x_{2}$ &
  $x_{2}$ &
  $x_{2}$ &
  $x_{2}$ \\ \hline
$h_{6}$ &
  8 &
  8 &
  $x_{3}$ &
  $x_{3}$ &
  $x_{3}$ &
  $x_{3}$ &
  $x_{3}$ &
  $x_{3}$ &
  $x_{3}$ &
  $x_{3}$ &
  $x_{3}$ &
  $x_{3}$ &
  $x_{3}$ &
  $x_{3}$ &
  $x_{3}$ \\ \hline
$z_{7}$ &
  0 &
  0 &
  0 &
  $x_{4}$ &
  $x_{4}$ &
  $x_{4}$ &
  $x_{4}$ &
  $x_{4}$ &
  $x_{4}$ &
  $x_{4}$ &
  $x_{4}$ &
  $x_{4}$ &
  $x_{4}$ &
  $x_{4}$ &
  $x_{4}$ \\ \hline
$\varepsilon_{7}$ &
  0 &
  0 &
  0 &
  0 &
  $x_{5}$ &
  $x_{5}$ &
  $x_{5}$ &
  $x_{5}$ &
  $x_{5}$ &
  $x_{5}$ &
  $x_{5}$ &
  $x_{5}$ &
  $x_{5}$ &
  $x_{5}$ &
  $x_{5}$ \\ \hline
$h_{7}$ &
  8 &
  8 &
  8 &
  8 &
  8 &
  $x_{6}$ &
  $x_{6}$ &
  $x_{6}$ &
  $x_{6}$ &
  $x_{6}$ &
  $x_{6}$ &
  $x_{6}$ &
  $x_{6}$ &
  $x_{6}$ &
  $x_{6}$ \\ \hline
$z_{8}$ &
  0 &
  0 &
  0 &
  0 &
  0 &
  0 &
  $x_{7}$ &
  $x_{7}$ &
  $x_{7}$ &
  $x_{7}$ &
  $x_{7}$ &
  $x_{7}$ &
  $x_{7}$ &
  $x_{7}$ &
  $x_{7}$ \\ \hline
$\varepsilon_{8}$ &
  0 &
  0 &
  0 &
  0 &
  0 &
  0 &
  0 &
  $x_{8}$ &
  $x_{8}$ &
  $x_{8}$ &
  $x_{8}$ &
  $x_{8}$ &
  $x_{8}$ &
  $x_{8}$ &
  $x_{8}$ \\ \hline
$h_{8}$ &
  8 &
  8 &
  8 &
  8 &
  8 &
  8 &
  8 &
  8 &
  $x_{9}$ &
  $x_{9}$ &
  $x_{9}$ &
  $x_{9}$ &
  $x_{9}$ &
  $x_{9}$ &
  $x_{9}$ \\ \hline
$z_{9}$ &
  0 &
  0 &
  0 &
  0 &
  0 &
  0 &
  0 &
  0 &
  0 &
  $x_{10}$ &
  $x_{10}$ &
  $x_{10}$ &
  $x_{10}$ &
  $x_{10}$ &
  $x_{10}$ \\ \hline
$\varepsilon_{9}$ &
  0 &
  0 &
  0 &
  0 &
  0 &
  0 &
  0 &
  0 &
  0 &
  0 &
  $x_{11}$ &
  $x_{11}$ &
  $x_{11}$ &
  $x_{11}$ &
  $x_{11}$ \\ \hline
$h_{9}$ &
  8 &
  8 &
  8 &
  8 &
  8 &
  8 &
  8 &
  8 &
  8 &
  8 &
  8 &
  $x_{12}$ &
  $x_{12}$ &
  $x_{12}$ &
  $x_{12}$ \\ \hline
$z_{10}$ &
  0 &
  0 &
  0 &
  0 &
  0 &
  0 &
  0 &
  0 &
  0 &
  0 &
  0 &
  0 &
  $x_{13}$ &
  $x_{13}$ &
  $x_{13}$ \\ \hline
$\varepsilon_{10}$ &
  0 &
  0 &
  0 &
  0 &
  0 &
  0 &
  0 &
  0 &
  0 &
  0 &
  0 &
  0 &
  0 &
  $x_{14}$ &
  $x_{14}$ \\ \hline
$h_{10}$ &
  8 &
  8 &
  8 &
  8 &
  8 &
  8 &
  8 &
  8 &
  8 &
  8 &
  8 &
  8 &
  8 &
  8 &
  $x_{15}$ \\ \hline
\caption{Equivalence Matrix of trigger characterization with each design variable for $1\leq d \leq 15$. Units in mm.}
\label{tab:equiv_table_1}\\
\end{longtable}

%
\begin{longtable}[c]{|l|l|l|l|l|l|l|l|l|l|l|l|l|l|l|l|}
\hline
Variable/$d$ &
  16 &
  17 &
  18 &
  19 &
  20 &
  21 &
  22 &
  23 &
  24 &
  25 &
  26 &
  27 &
  28 &
  29 &
  30 \\ \hline \hline
\endfirsthead
\endhead
$z_{1}$ &
  $x_{1}$ &
  $x_{1}$ &
  $x_{1}$ &
  $x_{1}$ &
  $x_{1}$ &
  $x_{1}$ &
  $x_{1}$ &
  $x_{1}$ &
  $x_{1}$ &
  $x_{1}$ &
  $x_{1}$ &
  $x_{1}$ &
  $x_{1}$ &
  $x_{1}$ &
  $x_{1}$ \\ \hline
$\varepsilon_{1}$ &
  $x_{2}$ &
  $x_{2}$ &
  $x_{2}$ &
  $x_{2}$ &
  $x_{2}$ &
  $x_{2}$ &
  $x_{2}$ &
  $x_{2}$ &
  $x_{2}$ &
  $x_{2}$ &
  $x_{2}$ &
  $x_{2}$ &
  $x_{2}$ &
  $x_{2}$ &
  $x_{2}$ \\ \hline
$h_{1}$ &
  $x_{3}$ &
  $x_{3}$ &
  $x_{3}$ &
  $x_{3}$ &
  $x_{3}$ &
  $x_{3}$ &
  $x_{3}$ &
  $x_{3}$ &
  $x_{3}$ &
  $x_{3}$ &
  $x_{3}$ &
  $x_{3}$ &
  $x_{3}$ &
  $x_{3}$ &
  $x_{3}$ \\ \hline
$z_{2}$ &
  $x_{4}$ &
  $x_{4}$ &
  $x_{4}$ &
  $x_{4}$ &
  $x_{4}$ &
  $x_{4}$ &
  $x_{4}$ &
  $x_{4}$ &
  $x_{4}$ &
  $x_{4}$ &
  $x_{4}$ &
  $x_{4}$ &
  $x_{4}$ &
  $x_{4}$ &
  $x_{4}$ \\ \hline
$\varepsilon_{2}$ &
  $x_{5}$ &
  $x_{5}$ &
  $x_{5}$ &
  $x_{5}$ &
  $x_{5}$ &
  $x_{5}$ &
  $x_{5}$ &
  $x_{5}$ &
  $x_{5}$ &
  $x_{5}$ &
  $x_{5}$ &
  $x_{5}$ &
  $x_{5}$ &
  $x_{5}$ &
  $x_{5}$ \\ \hline
$h_{2}$ &
  $x_{6}$ &
  $x_{6}$ &
  $x_{6}$ &
  $x_{6}$ &
  $x_{6}$ &
  $x_{6}$ &
  $x_{6}$ &
  $x_{6}$ &
  $x_{6}$ &
  $x_{6}$ &
  $x_{6}$ &
  $x_{6}$ &
  $x_{6}$ &
  $x_{6}$ &
  $x_{6}$ \\ \hline
$z_{3}$ &
  $x_{7}$ &
  $x_{7}$ &
  $x_{7}$ &
  $x_{7}$ &
  $x_{7}$ &
  $x_{7}$ &
  $x_{7}$ &
  $x_{7}$ &
  $x_{7}$ &
  $x_{7}$ &
  $x_{7}$ &
  $x_{7}$ &
  $x_{7}$ &
  $x_{7}$ &
  $x_{7}$ \\ \hline
$\varepsilon_{3}$ &
  $x_{8}$ &
  $x_{8}$ &
  $x_{8}$ &
  $x_{8}$ &
  $x_{8}$ &
  $x_{8}$ &
  $x_{8}$ &
  $x_{8}$ &
  $x_{8}$ &
  $x_{8}$ &
  $x_{8}$ &
  $x_{8}$ &
  $x_{8}$ &
  $x_{8}$ &
  $x_{8}$ \\ \hline
$h_{3}$ &
  $x_{9}$ &
  $x_{9}$ &
  $x_{9}$ &
  $x_{9}$ &
  $x_{9}$ &
  $x_{9}$ &
  $x_{9}$ &
  $x_{9}$ &
  $x_{9}$ &
  $x_{9}$ &
  $x_{9}$ &
  $x_{9}$ &
  $x_{9}$ &
  $x_{9}$ &
  $x_{9}$ \\ \hline
$z_{4}$ &
  $x_{10}$ &
  $x_{10}$ &
  $x_{10}$ &
  $x_{10}$ &
  $x_{10}$ &
  $x_{10}$ &
  $x_{10}$ &
  $x_{10}$ &
  $x_{10}$ &
  $x_{10}$ &
  $x_{10}$ &
  $x_{10}$ &
  $x_{10}$ &
  $x_{10}$ &
  $x_{10}$ \\ \hline
$\varepsilon_{4}$ &
  $x_{11}$ &
  $x_{11}$ &
  $x_{11}$ &
  $x_{11}$ &
  $x_{11}$ &
  $x_{11}$ &
  $x_{11}$ &
  $x_{11}$ &
  $x_{11}$ &
  $x_{11}$ &
  $x_{11}$ &
  $x_{11}$ &
  $x_{11}$ &
  $x_{11}$ &
  $x_{11}$ \\ \hline
$h_{4}$ &
  $x_{12}$ &
  $x_{12}$ &
  $x_{12}$ &
  $x_{12}$ &
  $x_{12}$ &
  $x_{12}$ &
  $x_{12}$ &
  $x_{12}$ &
  $x_{12}$ &
  $x_{12}$ &
  $x_{12}$ &
  $x_{12}$ &
  $x_{12}$ &
  $x_{12}$ &
  $x_{12}$ \\ \hline
$z_{5}$ &
  $x_{13}$ &
  $x_{13}$ &
  $x_{13}$ &
  $x_{13}$ &
  $x_{13}$ &
  $x_{13}$ &
  $x_{13}$ &
  $x_{13}$ &
  $x_{13}$ &
  $x_{13}$ &
  $x_{13}$ &
  $x_{13}$ &
  $x_{13}$ &
  $x_{13}$ &
  $x_{13}$ \\ \hline
$\varepsilon_{5}$ &
  $x_{14}$ &
  $x_{14}$ &
  $x_{14}$ &
  $x_{14}$ &
  $x_{14}$ &
  $x_{14}$ &
  $x_{14}$ &
  $x_{14}$ &
  $x_{14}$ &
  $x_{14}$ &
  $x_{14}$ &
  $x_{14}$ &
  $x_{14}$ &
  $x_{14}$ &
  $x_{14}$ \\ \hline
$h_{5}$ &
  $x_{15}$ &
  $x_{15}$ &
  $x_{15}$ &
  $x_{15}$ &
  $x_{15}$ &
  $x_{15}$ &
  $x_{15}$ &
  $x_{15}$ &
  $x_{15}$ &
  $x_{15}$ &
  $x_{15}$ &
  $x_{15}$ &
  $x_{15}$ &
  $x_{15}$ &
  $x_{15}$ \\ \hline
$z_{6}$ &
  $x_{16}$ &
  $x_{16}$ &
  $x_{16}$ &
  $x_{16}$ &
  $x_{16}$ &
  $x_{16}$ &
  $x_{16}$ &
  $x_{16}$ &
  $x_{16}$ &
  $x_{16}$ &
  $x_{16}$ &
  $x_{16}$ &
  $x_{16}$ &
  $x_{16}$ &
  $x_{16}$ \\ \hline
$\varepsilon_{6}$ &
  0 &
  $x_{17}$ &
  $x_{17}$ &
  $x_{17}$ &
  $x_{17}$ &
  $x_{17}$ &
  $x_{17}$ &
  $x_{17}$ &
  $x_{17}$ &
  $x_{17}$ &
  $x_{17}$ &
  $x_{17}$ &
  $x_{17}$ &
  $x_{17}$ &
  $x_{17}$ \\ \hline
$h_{6}$ &
  8 &
  8 &
  $x_{18}$ &
  $x_{18}$ &
  $x_{18}$ &
  $x_{18}$ &
  $x_{18}$ &
  $x_{18}$ &
  $x_{18}$ &
  $x_{18}$ &
  $x_{18}$ &
  $x_{18}$ &
  $x_{18}$ &
  $x_{18}$ &
  $x_{18}$ \\ \hline
$z_{7}$ &
  0 &
  0 &
  0 &
  $x_{19}$ &
  $x_{19}$ &
  $x_{19}$ &
  $x_{19}$ &
  $x_{19}$ &
  $x_{19}$ &
  $x_{19}$ &
  $x_{19}$ &
  $x_{19}$ &
  $x_{19}$ &
  $x_{19}$ &
  $x_{19}$ \\ \hline
$\varepsilon_{7}$ &
  0 &
  0 &
  0 &
  0 &
  $x_{20}$ &
  $x_{20}$ &
  $x_{20}$ &
  $x_{20}$ &
  $x_{20}$ &
  $x_{20}$ &
  $x_{20}$ &
  $x_{20}$ &
  $x_{20}$ &
  $x_{20}$ &
  $x_{20}$ \\ \hline
$h_{7}$ &
  8 &
  8 &
  8 &
  8 &
  8 &
  $x_{21}$ &
  $x_{21}$ &
  $x_{21}$ &
  $x_{21}$ &
  $x_{21}$ &
  $x_{21}$ &
  $x_{21}$ &
  $x_{21}$ &
  $x_{21}$ &
  $x_{21}$ \\ \hline
$z_{8}$ &
  0 &
  0 &
  0 &
  0 &
  0 &
  0 &
  $x_{22}$ &
  $x_{22}$ &
  $x_{22}$ &
  $x_{22}$ &
  $x_{22}$ &
  $x_{22}$ &
  $x_{22}$ &
  $x_{22}$ &
  $x_{22}$ \\ \hline
$\varepsilon_{8}$ &
  0 &
  0 &
  0 &
  0 &
  0 &
  0 &
  0 &
  $x_{23}$ &
  $x_{23}$ &
  $x_{23}$ &
  $x_{23}$ &
  $x_{23}$ &
  $x_{23}$ &
  $x_{23}$ &
  $x_{23}$ \\ \hline
$h_{8}$ &
  8 &
  8 &
  8 &
  8 &
  8 &
  8 &
  8 &
  8 &
  $x_{24}$ &
  $x_{24}$ &
  $x_{24}$ &
  $x_{24}$ &
  $x_{24}$ &
  $x_{24}$ &
  $x_{24}$ \\ \hline
$z_{9}$ &
  0 &
  0 &
  0 &
  0 &
  0 &
  0 &
  0 &
  0 &
  0 &
  $x_{25}$ &
  $x_{25}$ &
  $x_{25}$ &
  $x_{25}$ &
  $x_{25}$ &
  $x_{25}$ \\ \hline
$\varepsilon_{9}$ &
  0 &
  0 &
  0 &
  0 &
  0 &
  0 &
  0 &
  0 &
  0 &
  0 &
  $x_{26}$ &
  $x_{26}$ &
  $x_{26}$ &
  $x_{26}$ &
  $x_{26}$ \\ \hline
$h_{9}$ &
  8 &
  8 &
  8 &
  8 &
  8 &
  8 &
  8 &
  8 &
  8 &
  8 &
  8 &
  $x_{27}$ &
  $x_{27}$ &
  $x_{27}$ &
  $x_{27}$ \\ \hline
$z_{10}$ &
  0 &
  0 &
  0 &
  0 &
  0 &
  0 &
  0 &
  0 &
  0 &
  0 &
  0 &
  0 &
  $x_{28}$ &
  $x_{28}$ &
  $x_{28}$ \\ \hline
$\varepsilon_{10}$ &
  0 &
  0 &
  0 &
  0 &
  0 &
  0 &
  0 &
  0 &
  0 &
  0 &
  0 &
  0 &
  0 &
  $x_{29}$ &
  $x_{29}$ \\ \hline
$h_{10}$ &
  8 &
  8 &
  8 &
  8 &
  8 &
  8 &
  8 &
  8 &
  8 &
  8 &
  8 &
  8 &
  8 &
  8 &
  $x_{30}$ \\ \hline
\caption{Equivalence Matrix of trigger characterization with each design variable for $16\leq d \leq 30$. Units in mm.}
\label{tab:equiv_table_2}\\
\end{longtable}

\end{document}